\documentclass{article}

\usepackage[final]{neurips_2024}

\usepackage[utf8]{inputenc} % allow utf-8 input
\usepackage[T1]{fontenc}    % use 8-bit T1 fonts
\usepackage{hyperref}       % hyperlinks
\usepackage{url}            % simple URL typesetting
\usepackage{booktabs}       % professional-quality tables
\usepackage{amsfonts}       % blackboard math symbols
\usepackage{nicefrac}       % compact symbols for 1/2, etc.
\usepackage{microtype}      % microtypography
\usepackage{xcolor}         % colors

\usepackage{makecell} % 表格内容居中

\usepackage{booktabs} % 表格边框
\usepackage{array}
\usepackage{multirow}
\usepackage{xcolor}
\usepackage{hyperref}  % 引用章节索引
\usepackage{stfloats}  % 强制双栏图片在本页
\usepackage{amssymb}
\usepackage{lineno}
\usepackage{graphicx}
\usepackage[export]{adjustbox}
\usepackage{times}
\usepackage{tabu}
\usepackage{caption}
\usepackage{setspace}
\usepackage{amsmath}
\usepackage{lineno}
\usepackage{soul}
\usepackage{enumitem}  % 调整列表间距
\usepackage{mdwlist}  % 调整 item 间距
\usepackage{bbding}  % 对号错号
\usepackage{pifont} %\ding 
\usepackage{wrapfig, lipsum}
\usepackage{CJKutf8}

\definecolor{LimeGreen}{rgb}{0.2, 0.8, 0.2}
\definecolor{Red}{rgb}{1, 0, 0}

\title{Towards Comprehensive Detection of Chinese Harmful Memes}

\author{Junyu Lu\textsuperscript{1}, Bo Xu\textsuperscript{1}, Xiaokun Zhang\textsuperscript{1}, Hongbo Wang\textsuperscript{1}, \\
        \textbf{Haohao Zhu\textsuperscript{1}, Dongyu Zhang\textsuperscript{2},  Liang Yang\textsuperscript{1,3}, Hongfei Lin\textsuperscript{1}} \\
        \textsuperscript{1} School of Computer Science and Technology, \textsuperscript{2} School of Foreign Languages, 
        \\\textsuperscript{3} Key Laboratory of Social Computing and Cognitive Intelligence，\\
        Dalian University of Technology, China \\ 
        \texttt{dutljy,zhuhh@mail.dlut.edu.cn,  dawnkun1993,1846742523a@gmail.com}\\
        \texttt{xubo,zhangdongyu,liang,hflin@dlut.edu.cn}}

\begin{document}
\begin{CJK*}{UTF8}{gbsn}

\maketitle

\begin{abstract}
% It is imperative to incorporate diverse types of Chinese harmful memes into dataset construction, as well as to integrate the contextual information of meme content for detection.

Harmful memes have proliferated on the Chinese Internet, while research on detecting Chinese harmful memes significantly lags behind due to the absence of reliable datasets and effective detectors.
To this end, we focus on the comprehensive detection of Chinese harmful memes.
We construct \textsc{ToxiCN MM}, the first Chinese harmful meme dataset, which consists of 12,000 samples with fine-grained annotations for various meme types. 
Additionally, we propose a baseline detector, Multimodal Knowledge Enhancement (MKE), incorporating contextual information of meme content generated by the LLM to enhance the understanding of Chinese memes.
During the evaluation phase, we conduct extensive quantitative experiments and qualitative analyses on multiple baselines, including LLMs and our MKE. 
The experimental results indicate that detecting Chinese harmful memes is challenging for existing models while demonstrating the effectiveness of MKE.\footnote{Resources of this paper are available at \url{https://github.com/DUT-lujunyu/ToxiCN_MM}.}

% We first define "\textit{Chinese harmful memes}" to adapt to the Chinese online environment. 

\textit{\textbf{Disclaimer}: The samples presented by this paper may be considered profane, offensive, or vulgar.}

\end{abstract}

\section{Introduction}

With the development of the Internet, harmful memes on the web have become increasingly rampant. 
% To circumvent text moderation mechanisms of the platforms, harmful memes emerge as an alternative form of expression.
Harmful memes are typically defined as multimodal units consisting of an image and embedded text that cause harm to an individual, an organization, a community, or a social group by specifically targeting social entities \citep{DBLP:conf/acl/PramanickDMSANC21, DBLP:conf/ijcai/SharmaAADMFHSN022}.
They may exacerbate social divisions, trigger discriminatory behaviors, and harm social harmony and unity \citep{DBLP:conf/nips/KielaFMGSRT20}.
Due to their negative impact on society, the widespread dissemination of harmful memes has been widely recognized as a growing concern.

In recent years, researchers have made substantial progress in detecting harmful memes.
Several datasets, including HMC \citep{DBLP:conf/nips/KielaFMGSRT20}, MMHS \citep{DBLP:conf/wacv/GomezGGK20}, Harm-C, and Harm-P \citep{DBLP:conf/acl/PramanickDMSANC21}, have been established, and various detectors have been proposed \cite{DBLP:conf/www/HeeLC22, DBLP:conf/www/AggarwalCDSMZ023, DBLP:conf/emnlp/PramanickSDAN021, DBLP:conf/naacl/SharmaAN022, DBLP:conf/emnlp/CaoLC022}.
However, most existing studies only focus on English memes.
In contrast, Chinese harmful meme detection remains largely unexplored, presenting challenges in building reliable datasets and developing effective detectors.

% which do not adapt to the Chinese online environment, 
% However, most of these studies only focus on English memes. 
% In contrast, the detection of Chinese harmful memes lags significantly, lacking reliable benchmark datasets and effective detectors. 
% Given the cultural difference, the development of appropriate Chinese datasets and detectors becomes imperative to adapt to the characteristics of Chinese harmful memes.
% Due to the difference of cultural background, it is necessary to construct corresponding datasets and detectors for Chinese.

% \citep{jahan2021systematic}. 

\begin{figure*}[htpb]
\centering
\includegraphics[width=14cm]{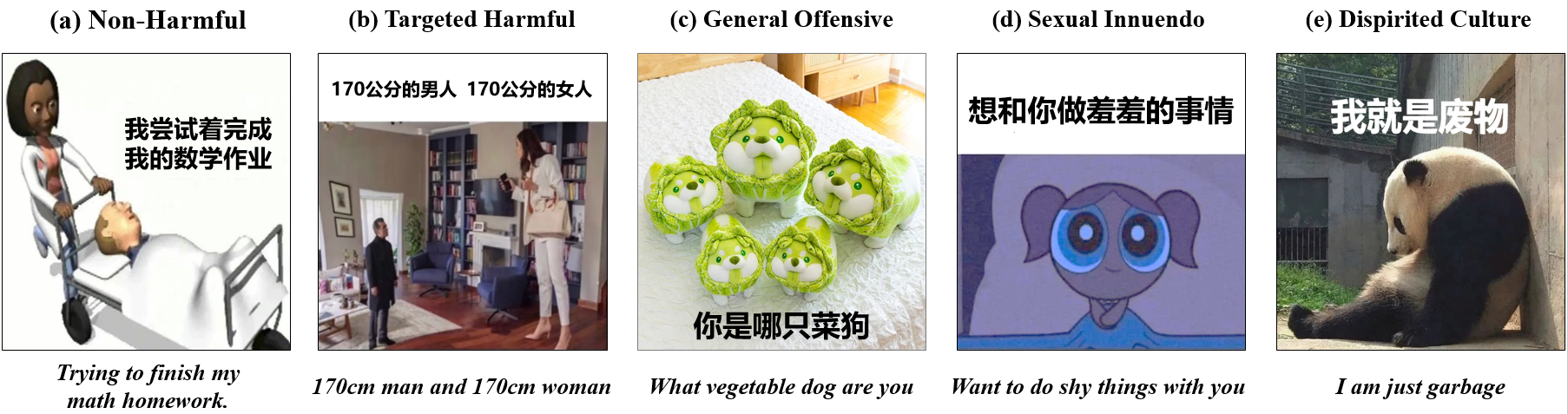}
% \vspace{-0.25in}
\caption{Illustration of memes in Chinese. (a) is a harmless meme that humorously expresses concern about math homework.
(b) is a targeted harmful meme conveying gender bias through the height differences between the man and woman. 
(c) contains general offense without specific targets, where "\textit{vegetable dog}" is a Chinese insult, implying incompetent people. 
(d) subtly conveys sexual innuendo with an idiom "\textit{shy things}" (alluding to "\textit{sexual intercourse}").
(e) spreads dispirited culture by comparing oneself to  "\textit{garbage}".}
\vspace{0.1in}
\label{introduction}
\end{figure*}

% In terms of harmful types, 
On one hand, the types of Chinese harmful memes are diverse.
In addition to those targeting specific social entities, many memes on Chinese platforms contain general offense, sexual innuendo, or dispirited culture \citep{liu2016}, as shown in Figure~\ref{introduction}. 
Despite lacking specific targets, they still exhibit potential toxicity, subtly propagating negative values that could lead to serious consequences like violent acts and sexual harassment \citep{lin2019}.
To adapt to the online environment, it is crucial to consider this diversity when constructing the dataset.

On the other hand, understanding the semantics of Chinese harmful memes presents a significant challenge for detectors, necessitating contextual information from both textual and visual elements.
For example, Exp. (a) of Figure~\ref{introduction} illustrates gender bias through the height difference between the man and the woman in the image.
In contrast, the inline text of Exp. (b) introduces a Chinese insult, \textit{"vegetable dog"}, to tease others, which implies incompetent people.
Therefore, incorporating such information is necessary for the effective detection of Chinese harmful memes.

In this paper, we facilitate the detection of Chinese harmful memes, primarily focusing on two aspects: \textbf{dataset construction} and \textbf{detector development}.
For the dataset construction, we first propose the definition of "\textit{Chinese harmful memes}" as guidance, accurately adapting to the Chinese online environment. 
Based on the definition, we focus on both targeted harmful memes and those exhibiting potential toxicity without specific targets. 
We conduct fine-grained annotation for harmful memes collected from Chinese online platforms, analyzing their harmful types and combination features of textual and visual information.
\textsc{ToxiCN MM} dataset is then constructed, encompassing 12,000 samples containing different harmful types.
Two progressive tasks are established using the dataset: (I) Detect if a meme is harmful, and (II) If harmful, further identify its harmful type.

For the detector development, we present a Multimodal Knowledge Enhancement (MKE) detector, integrating the contextual information of meme content for effective detection. 
We first utilize the large language model (LLM) to capture the context of both the text and image of the meme, leveraging its extensive knowledge acquired through pre-training \cite{zhao2023survey}.  
This information is then integrated into a trainable detector as enhanced captions to improve the understanding of memes.
In the experimental phase, we evaluate the detection performance of various baselines, including both traditional pre-trained language models (PLMs) and large language models (LLMs), providing a benchmark for evaluation.
Experimental results show the effectiveness of our MKE.
The main contributions of this paper are summarized as follows:

% \vspace{-0.05in}
% 
\begin{itemize}[itemsep=0.5pt]
    % \item We propose the linguistic definition of "\textit{Chinese harmful memes}" according to their characteristic, adapting to the Chinese online environment. We not only focus on targeted harmful memes but also on those that exhibit potential toxicity without specific targets. 
    \item We construct, to the best of our knowledge, the first Chinese harmful meme dataset \textsc{ToxiCN MM}, which comprises 12,000 diverse samples. We conduct a fine-grained annotation to analyze their harmful types and modality combination features.
    \item We present Multimodal Knowledge Enhancement (MKE) as a baseline detector, integrating contextual information of meme content generated by the LLM to improve the detector's understanding of Chinese harmful memes.
    \item We utilize \textsc{ToxiCN MM} as a benchmark to evaluate the detection performance of various baseline models. Extensive quantitative experiments and qualitative analysis illustrate the effect of MKE. We summarize the challenges of Chinese harmful memes detection.
\end{itemize}

% enhancing its understanding of memes, 

\section{Related Work}
% \textbf{Linguistic Features of Hate Speech.} Some work has examined the connection between hate speech and other associated toxic comments, especially general offensive language \citep{djuric2015hate, DBLP:conf/icwsm/DavidsonWMW17, DBLP:conf/aaai/MathewSYBG021}. There are also many researchers exploring the expression of hate speech dissemination. \citet{dardenne2007insidious, djuric2015hate, DBLP:journals/csur/FortunaN18, ibrohim2019multi} pointed out that, in addition to directly insulting and defaming minority groups, some implicit bias, such as stereotyping and microaggression, should also be considered as hate speech. Besides that, \citet{DBLP:conf/lrec/ChirilMBMOC20} proposed that reporting of hate experience breeds social bias and harms minorities, even although the publisher is well-intentioned. 

\begin{table*}
  \centering
\small
\renewcommand{\arraystretch}{1.2}
    \begin{tabular}{m{5.5cm}m{1.5cm}m{1.4cm}m{1.3cm}cc<{\centering}}%{llllll}
    \bottomrule
    \textbf{Work}  & \textbf{Language} & \multicolumn{1}{l}{\textbf{Size}} & \multicolumn{1}{l}{\textbf{Ratio}} & \multicolumn{1}{l}{\textbf{Pot. Tox.}} \\ \hline 
    HarMeme \citep{DBLP:conf/acl/PramanickDMSANC21} & English &  3,544 & 34.96\% & \XSolidBrush \\ 
    Harm-C \citep{DBLP:conf/emnlp/PramanickSDAN021} & English &  3,013 & 35.31\% & \XSolidBrush \\ 
    Harm-P \citep{DBLP:conf/emnlp/PramanickSDAN021} & English &  3,020 & 49.21\% & \XSolidBrush \\  
    HMC \citep{DBLP:conf/nips/KielaFMGSRT20}   & English & 10,000 & 38.00\% & \XSolidBrush \\  
    MMHS150K \citep{DBLP:conf/wacv/GomezGGK20} & English & 150,000 & 24.68\% & \XSolidBrush \\  
    TamilMemes \citep{suryawanshi-etal-2020-dataset} & Indian & 2,969 & 65.71\% & \XSolidBrush \\ 
    MUTE  \citep{DBLP:conf/ijcnlp/HossainSH22} & Bengali & 4,158 & 37.87\% & \XSolidBrush \\  
    MAMI \citep{DBLP:conf/semeval/FersiniGRSCRLS22}  & English & 11,000 & 50.00\% & \XSolidBrush \\ 
    % HatRed \citep{DBLP:conf/ijcai/HeeCL23}  & English & 3,228  & 100.00\% & \XSolidBrush & \Checkmark \\ \hline 
    \hline
    \textsc{ToxiCN MM} (ours) & Chinese & 12,000   & 31.89\%  & \Checkmark \\ 
    \toprule
    \end{tabular}% 
    % \vspace{-0.1in}
  \caption{Information of harmful meme datasets, in terms of \textit{Language}, \textit{Size}, \textit{Ratio} of harmful samples, and whether containing memes exhibiting potential toxicity (\textit{Pot. Tox.}) without specific targets.}
  % \vspace{-0.125in}
  \label{dataset_work}%
\end{table*}%

\textbf{Harmful Meme.} 
In recent years, researchers have noticed the importance of detecting harmful memes. 
Several datasets have been established \cite{DBLP:conf/nips/KielaFMGSRT20, DBLP:conf/wacv/GomezGGK20, DBLP:conf/acl/PramanickDMSANC21}.
Nevertheless, most current studies only focus on English, while research on detecting Chinese harmful memes remains unexplored.
To this end, we present the first Chinese harmful meme dataset \textsc{ToxiCN MM}.
Here we list \tablename~\ref{dataset_work} to compare existing datasets with \textsc{ToxiCN MM}.

Existing studies define “\textit{harmful memes}” as memes that cause harm to specific social entities, based on their social attributes such as religion, race, and gender \citep{DBLP:conf/acl/PramanickDMSANC21}. 
However, this definition does not fully apply to the Chinese Internet, where harmful memes often exhibit potential toxicity without specific targets but still perpetuate negative cultural values \citep{liu2016, zhang2021}.
In this paper, we consider different harmful types for comprehensive detection.

Several methods have been proposed for harmful meme detection, primarily focusing on modeling based on targets of memes \cite{DBLP:conf/emnlp/PramanickSDAN021, DBLP:conf/naacl/SharmaAN022, DBLP:conf/mir/KoutlisSP23, DBLP:conf/www/JiRN23}. However, they do not apply to Chinese harmful memes, which often lack specific targets.
Despite this limitation, some methods enhance the detector's understanding of memes by integrating image descriptions to make decisions \cite{DBLP:conf/emnlp/BlaierMW21, DBLP:conf/emnlp/CaoLC022}. 
These studies still inspire us to integrate more comprehensive contextual information of meme content into detectors.

% , leaving a gap in the study of detecting such content.
% In this paper, we define \textit{"Chinese harmful memes"} to encompass these memes.

% Despite this limitation, some methods enhance the detector's understanding of memes by integrating image descriptions to make decisions \citep{DBLP:conf/emnlp/BlaierMW21, DBLP:conf/emnlp/CaoLC022}. 
% These studies still inspire us to integrate more comprehensive contextual information of meme content into detectors.

% Nevertheless, to enhance understanding of memes, researchers have integrated image descriptions using caption generators \citep{DBLP:conf/emnlp/BlaierMW21, DBLP:conf/emnlp/CaoLC022}. 
% Inspired by this, we employ the LLM to capture the contextual information of meme content from both inline text and images as captions input into the detector.

% Furthermore, existing studies lack sufficient analysis of the common and unique features of harmful memes in different languages. 

\textbf{Toxic Language.} 
Harmful memes are closely linked to toxic language \citep{DBLP:conf/nips/KielaFMGSRT20}, which are rude, disrespectful, or unreasonable, and can drive people away from conversations \citep{DBLP:conf/aies/DixonLSTV18}. Chinese harmful memes frequently feature toxic language in their inline text, utilizing slang and linguistic phenomena like homophony.  
Therefore, understanding Chinese memes necessitates the incorporation of linguistic knowledge.

Additionally, labeling both harmful memes and toxic language is often subjective. 
Several studies have addressed this issue by focusing on mitigating the subjective bias of annotators during the construction of toxic language datasets \citep{DBLP:conf/naacl/WaseemH16, ross2017measuring, DBLP:conf/acl/ZeinertID20, DBLP:conf/emnlp/FortunaDWT22, DBLP:conf/acl/LuXZMYL23, DBLP:conf/nlpcc/WangLLYXL23}, enhancing the reliability of datasets.
% Many measures have been implemented to mitigate the subjective bias of the annotators. 
In this paper, we adopt these measures as a reference in the annotation process of our \textsc{ToxiCN MM}.

\section{Dataset Construction}
\subsection{Overview}

\begin{figure*}
\centering
\includegraphics[width=14cm]{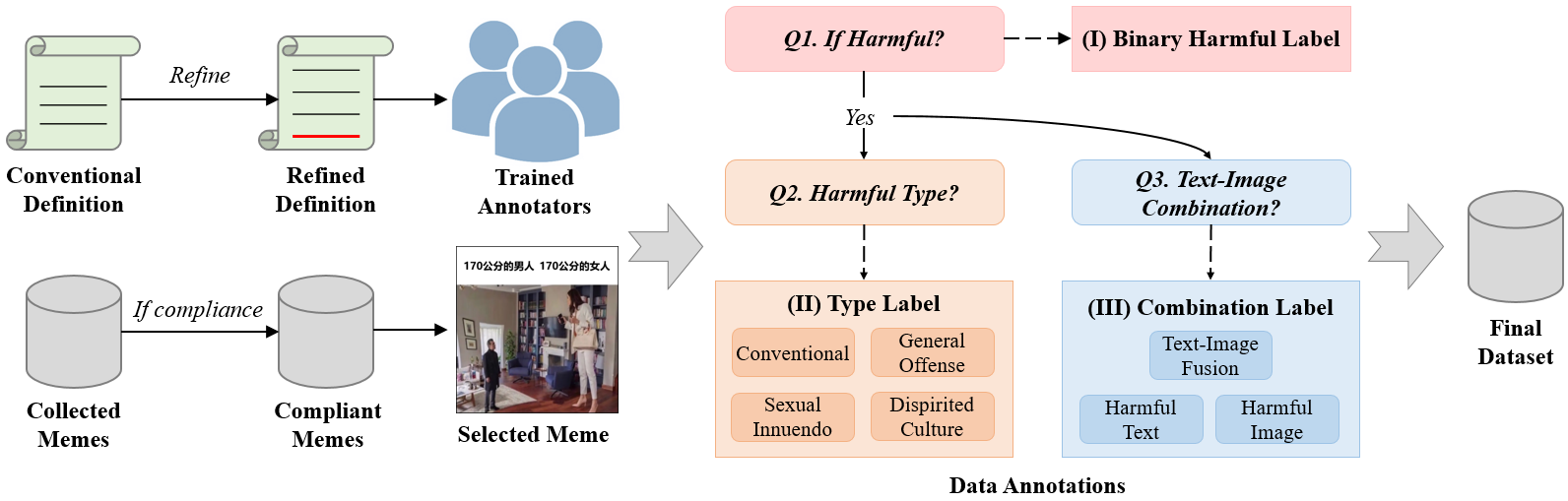}
% \vspace{-0.25in}
\caption{Illustration of \textsc{ToxiCN MM} construction procedure. According to the definition (Section \ref{Definition}), data collection and filtering (Section \ref{Collect}) and fine-grained annotations (Section \ref{Annotate}) are conducted sequentially. The final statistics of \textsc{ToxiCN MM} are presented in Section \ref{Statistics}.}
% \vspace{-0.1in}
\label{frame}
\end{figure*}

In this section, we detail the construction of our \textsc{ToxiCN MM} dataset. 
We first define "Chinese harmful memes" to guide the dataset annotation.
We then conduct fine-grained annotation for memes collected from Chinese platforms.
In addition to the basic binary labels, we analyze harmful memes from both harmful types and combinations of inline text and image information.
After the annotation, statistics of \textsc{ToxiCN MM} are presented. 
The diagram of dataset construction is shown in \figurename~\ref{frame}. 
 % Inter-Annotator Agreement (IAA) of each granularity and
% Finally, the Inter-Annotator Agreement (IAA) of each individual granularity and 
% the Inter-Annotator Agreement (IAA) of each individual granularity 
% After employing several measures to mitigate subjective biases,
% multiple data sources, including 
% , and an image retrieval engine \textit{Baidu Image} 
% to ensure the quality of the original data

\subsection{Definition Development}\label{Definition} 

The recognized definition of "\textit{harmful meme}" typically pertains to memes that target specific social entities.
However, numerous memes on the Chinese Internet diverge from this definition by only propagating negative values without specific targets, which can be equally detrimental to society. 
To adapt to the Chinese online environment, a refined definition is necessary. 
Here we introduce the definition of Chinese harmful memes:
% Despite this broader scope, these memes
% However, this definition falls short in addressing harmful memes that propagate negative values without specific targets. 

\textit{Chinese harmful memes are multimodal units consisting of an image and Chinese inline text that have the potential to cause harm to an individual, an organization, a community, a social group, or society as a whole. 
These memes can range from offense or joking that perpetuate harmful stereotypes towards specific social entities, to memes that are more subtle and general but still have the potential to cause harm. 
It is important to note that Chinese harmful memes can be created and spread intentionally or unintentionally. They often reflect and reinforce underlying negative values and cultural attitudes on the Chinese Internet, which are detrimental from legal or moral perspectives.}

According to the definition, we further identified the most common harmful types of memes on Chinese platforms based on the consensus of social psychology \citep{liu2016, lin2019} and communication \citep{peng2019, zheng2016} studies.
Specifically, it mainly includes \textit{targeted harmful}, \textit{general offense}, \textit{sexual innuendo}, and \textit{dispirited culture}.
The harm of these memes to individuals and society has been widely discussed.
In this study, we focus on these harmful types when constructing the dataset.
%  (see Appendix \ref{background})
% to balance the actual situation of the Chinese platform and the diversity of samples. To make the harmful samples in the dataset more representative,

% According to the definition, our research focuses not only on targeted harmful memes with specific targets, but also on other harmful types exhibiting potential toxicity without specific targets, 
% These memes are rampant on Chinese platforms, accurately reflecting the characteristics of Chinese harmful memes .
% in the following dataset construction and detection, 

\subsection{Data Collection and Filtering}\label{Collect}

Data collection is the basic work of constructing datasets, and its breadth and quality greatly affect the subsequent research. 
To ensure a comprehensive dataset, we collect Chinese memes from two well-known public online platforms, \textit{Weibo} and \textit{Baidu Tieba}, both widely representative of local users in China with active meme communities.
% Due to the content review mechanism of the platforms, harmful content is significantly limited \citep{deng-etal-2022-cold}. 
We first conduct a random crawl to obtain a diverse set of memes. 
To maximize the inclusion of harmful memes, we further focus our data crawl on sensitive topics commonly debated online (\textit{e.g.}, "\textit{gender}" and "\textit{region}").
Additionally, we also target memes expressing negative emotions and attitudes (\textit{e.g.}, "\textit{crazy}" and "\textit{Dispirited Culture}") to enrich the dataset with samples potentially exhibiting toxicity.
A total of about 14k memes are collected.
We then de-duplicate the data and filter out dirty samples including unreadable memes. 
The final dataset contains 12k refined memes.

Subsequently, we utilize \texttt{Baidu-OCR} to extract inline text from memes, which offers a high-precision service for Chinese text recognition.
To further enhance the sample quality, we also introduce a manual review process to examine the accuracy of the extracted text.
Specifically, we normalize the text by adding appropriate separators and removing additional line breaks and spaces.

% % To mitigate uncertainty in the process of data annotation. I
% [5pt]

\begin{wraptable}{r}{6cm}
\center
\small
% {ccccc}  <{\centering}
\scalebox{0.9}{
\begin{tabular}{cc} 
\toprule 
\textbf{Characteristic} & \textbf{Demographics}\\  
\midrule 
Gender & Male: 5, Female: 5    \\ 
Age & <20: 3, 20\textasciitilde30: 5, >30: 2 \\  
Race \& Region & Asians: 7, Others: 3  \\
 % Race &  From 5 different provinces \\
Education & UG: 3, PG: 4, PhD: 3  \\
\bottomrule
\end{tabular}
}
\vspace{0.05in}
\caption{Annotators demographics.}
\label{demographics}%
\end{wraptable}

\subsection{Data Annotation}\label{Annotate}

\subsubsection{Annotator Selection and Training}
Before the formal annotation process, it is crucial to select annotators carefully and mitigate their subjective bias, as this can significantly impact the quality of the dataset \citep{DBLP:conf/naacl/WaseemH16}.
To this end, we adopt the following measures: 
The majority of active users on Chinese platforms are between 12 and 35 years old. 
Considering Chinese laws that restrict individuals under 18 from engaging in activities that could harm their physical or mental health, we selected annotators aged 18 to 35. 
We assessed the annotators’ proficiency in Chinese meme culture through questionnaires and ensured diversity in terms of gender, region, and education level to enhance reproducibility.
The demographics of annotators are shown in \tablename~\ref{demographics}. 

During the training of annotators, we provided definitions of Chinese harmful memes, their various types, and conducted case analyses with diverse examples. 
To evaluate the annotators' abilities, we introduced three test groups of 100 memes each. 
Annotators labeled the memes independently, with researchers finalizing the labels. 
Post-round discussions were held to reduce errors, and detailed criteria, including edge cases, were established. 
Annotators improved from 63\% accuracy in the first round to 78\% in the final round, demonstrating the effectiveness of the training.

% This is where the table goes with text wrapping around it. You may 
% embed tabular environment inside wraptable environment and customize as you like.
% %------------------------------------------
% \begin{wraptable}{r}{5.5cm}
% \begin{tabular}{ccc}\\\toprule  
% Header-1 & Header-1 & Header-1 \\\midrule
% 2 &3 & 5\\  \midrule
% 2 &3 & 5\\  \midrule
% 2 &3 & 5\\  \bottomrule
% \end{tabular}
% \caption{A wrapped table going nicely inside the text.}\label{wrap-tab:1}
% \end{wraptable} 
% %------------------------------------------
% {\lipsum[2] 
% \par
% Table~\ref{wrap-tab:1} is a wrapped table.

\subsubsection{Label Annotation}

To guarantee the consistency of label annotation, we establish a comprehensive annotation framework as a guideline.
The specific process includes the following three stages.

%  based on the extended definition of "\textit{harmful memes}"

% In addition, if the sample is a recognized harmful meme, \textit{i.e.} it attacks specific targets, the targets are also annotated. 

\textbf{Whether Harmful}. 
The foundation of the labeling is to determine whether a meme is harmful or benign, which is a binary annotation.
We strictly follow the definition of "\textit{Chinese harmful memes}", focusing not only on targeted harmful memes but also on samples exhibiting potential toxicity without specific targets.

\textbf{Harmful Type}. 
In the second stage, we further refine the categorization of harmful memes, including targeted harmful, general offensive, sexual innuendo, and dispirited culture. 
% These types of harmful memes are rampant on Chinese platforms \citep{liu2016, lin2019}.
The annotation criteria for each harmful type are provided below.
\textbf{Targeted Harmful} memes express disgust, prejudice, or stereotypes towards specific individuals or social groups.  
In contrast, \textbf{General Offensive} memes encompass sarcastic or rude content but lack specific targets.
We also adhere to psychological and sociological definitions to classify the other two types:
\textbf{Sexual Innuendo} refers to memes that imply sexual intent to provoke sexual arousal \citep{bell1997innuendo}.
Here we label memes that contain suggestive elements but not sexism or sexual assault as such samples to distinguish them from targeted harmful memes.
And \textbf{Dispirited Culture} is characterized by the integration of decadent and desperate emotions, conveying a self-negative attitude \citep{QSNY201703001}.

\textbf{Modality Combination}. 
As multimodal units, harmful memes consist of both textual and visual modality, expressing toxicity through fused or independent features \citep{DBLP:conf/nips/KielaFMGSRT20}. 
To gain a more comprehensive understanding of how harmful content is propagated via memes, we classify them based on the toxic manifestation of these two modalities, exploring their individual and combined effects. 
Among them, \textbf{Text-Image Fusion} memes exhibit toxicity only through the combined effect of both modalities, while the text and image separately remain benign.
In contrast, \textbf{Harmful Text} and \textbf{Harmful Image} categories refer to one modality (either the text or the image) that independently exhibits toxicity.
% , while in \textbf{Harmful Text \& Image} memes, both the image and text components are independently toxic.

% We utilize an open-source annotation tool Doccano\footnote{\url{https://github.com/doccano/doccano}} to facilitate the annotation process. 
During the label annotation, each meme is labeled by at least three annotators. And we use a majority vote to assign the final label. 
In addition, specific targets of targeted harmful memes are provided. 
We then discuss the Inter-Annotator Agreement (IAA) of each granularity, as shown in Appendix \ref{IAA}.

\subsection{Statistics Description}\label{Statistics}

\begin{table*}
\renewcommand{\arraystretch}{1.2}
\small
  \centering
  \scalebox{0.9}{
    \begin{tabular}{m{1cm}<{\centering}|m{1.1cm}<{\centering}|m{1.1cm}<{\centering}|m{1cm}<{\centering}m{1cm}<{\centering}m{1cm}<{\centering}m{0.8cm}<{\centering}|m{0.8cm}<{\centering}m{1cm}<{\centering}m{1cm}<{\centering}|m{1cm}<{\centering}}
    %{c|c|c|cccc|ccc|c}
    \bottomrule
    \multirow{2}[4]{*}[5pt]{\textbf{Split}} & \multirow{2}[4]{*}[5pt]{\textbf{N-Harm.}} & \multirow{2}[4]{*}[5pt]{\textbf{Harm.}} & \multicolumn{4}{c|}{\textbf{Harmful Type Category}} & \multicolumn{3}{c|}{\textbf{Combination Category}} & \multirow{2}[4]{*}[5pt]{\textbf{Total}} \\
\cline{4-10}          &       &       & \textbf{Tg.} & \textbf{Off.} & \textbf{Sex.} & \textbf{Disp.} & \textbf{T-I} & \textbf{Harm.T} & \textbf{Harm.I} &  \\
    \hline
    \textbf{Train} & 6,538 & 3,062 & 813   & 1,198 & 731   & 320   & 1,082 & 1,754 & 276   & 9,600 \\
    \textbf{Test} & 1,635 & 765   & 203   & 300   & 183   & 79    & 271   & 438   & 69    & 2,400 \\
    \hline
    \textbf{Total} & 8,173 & 3,827 & 1,016 & 1,498 & 914   & 399   & 1,353 & 2,192 & 345   & 12,000 \\
    \toprule
    \end{tabular}%
    }
    % \vspace{-0.075in}
    \caption{Basic statistics of \textsc{ToxiCN MM}, listing the number of non-harmful (\textit{N-Harm.}) and harmful (\textit{Harm.}) samples, containing targeted harmful memes (\textit{Tg.}), general offense (\textit{Off.}), sexual innuendo (\textit{Sex.}), and dispirited culture (\textit{Disp.}), as well as
    each modality combination category (including text-image fusion (\textit{T-I}), harmful text (\textit{Harm.T}) and harmful image (\textit{Harm.I})).}   
    % \vspace{-0.12in}
  \label{statistics}%
\end{table*}%

For subsequent model training and evaluation, all samples in \textsc{ToxiCN MM} are divided into a training set and a test set at a ratio of 8:2, as detailed in \tablename~\ref{statistics}. 
We note that there exists a sample imbalance among different categories of harmful samples. 
Memes containing general offensive content (\textit{Off.}) constitute nearly 40\% of harmful memes. 
Regarding modality combinations, over 50\% of the inline text of harmful memes is harmful.
Given that the data distribution accurately reflects the real state of platforms, we do not introduce supplementary sampling to address existing imbalances.

\begin{wraptable}{r}{6.75cm}
\renewcommand{\arraystretch}{1.2}
  \centering
  \scalebox{0.9}{
  \small
    \begin{tabular}{m{1cm}<{\centering}|m{1cm}<{\centering}m{1cm}<{\centering}m{1cm}<{\centering}|m{1cm}<{\centering}}
    \toprule
          & \textbf{T-I} & \textbf{Harm.T} & \textbf{Harm.I} & \textbf{Total} \\
    \midrule
    \textbf{Tg.} & 575   & 404   & 47    & 1,016 \\
    \textbf{Off.} & 198   & 1,247 & 93    & 1,498 \\
    \textbf{Sex.} & 431   & 307   & 186   & 914 \\
    \textbf{Disp.} & 149   & 234   & 19    & 399 \\
    \midrule
    \textbf{Total} & 1,353 & 2,192 & 345   & 12,000 \\
    \bottomrule
    \end{tabular}%
    }
    % \vspace{0.05in}
    \caption{Modality combination distribution of different harmful types in \textsc{ToxiCN MM}.}
    % \vspace{-0.075in}
  \label{harmful_fusion}%
\end{wraptable}%

We then analyze the modality combinations across different harmful types, as shown in \tablename~\ref{harmful_fusion}. 
Each type displays distinct patterns in its modality combinations.
For example, memes containing general offense or dispirited culture (\textit{Disp.}) mainly feature inherently harmful inline text.
In contrast, over 50\% of targeted harmful memes (\textit{Conv.}) integrate multimodal features to express toxicity, where both text and image are individually benign. 
Moreover, there are 63 memes in which both text and image exhibit toxicity.

\section{Detector Development}

\subsection{Overview}

\begin{wrapfigure}{r}{6.75cm}
\centering
\includegraphics[width=7cm]{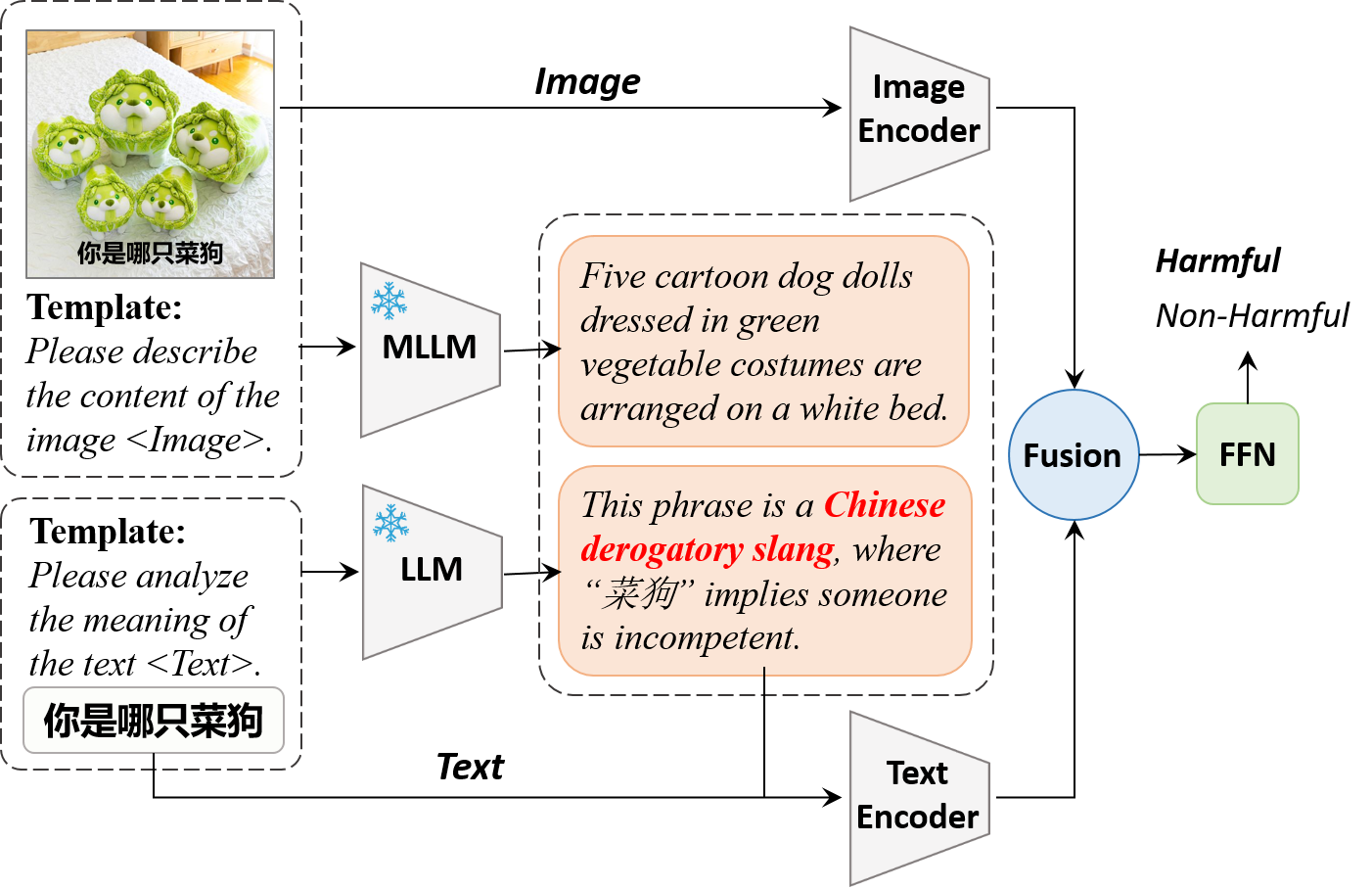}
\caption{Overview of MKE. The translation of the inline text is "what vegetable dog are you".}
\vspace{-0.1in}
\label{Toxic_small}
\end{wrapfigure}

To improve the detector's understanding of memes, we present a baseline detector, Multimodal Knowledge Enhancement (MKE), which integrates contextual information of meme content for more accurate predictions. 
We first leverage the LLM to capture the contextual context of memes and generate enhanced captions.  
Then, we fine-tune the detector by integrating the original inputs (i.e., text-image pairs) with the generated captions.
The illustration of MKE is shown in Figure~\ref{Toxic_small}.

% For a given meme, its inline text and image are encoded as $S\in\mathbb{R}^{d_s}$ and $V\in\mathbb{R}^{d_v}$, with $d_s$ and $d_v$ being the dimensions of the textual and visual vector spaces.

For a given meme, its inline text and image are encoded by modality-specific encoders, represented as $S \in \mathbb{R}^{d_s}$ for text and $V \in \mathbb{R}^{d_v}$ for the image, where $d_s$ and $d_v$ denote the dimensions of the textual and visual vector spaces, respectively.

\subsection{Knowledge Mining}

We instruct the LLM to generate enhanced captions for the meme by designing the instruction template, which respectively captures the contextual information from both the inline text and image.
To improve the understanding of the inline text that may contain slang, we enable LLM to incorporate language features unique to Chinese for semantic analysis.
The template is as follows:
"\textit{Considering Chinese linguistic characteristics, please analyze the meaning of the text <Text>.}" 
We further convert the image into textual descriptions with the multimodal large language model (MLLM), capturing harmful elements in the context of Chinese culture.
The template is designed as
"\textit{Considering Chinese cultural background, please describe the content of the image <Image>.}"

% that complement the inline text

In the process of knowledge mining, all parameters of LLMs are frozen.
To facilitate knowledge integration for subsequent detectors, the captions of inline text and images are represented by the text encoder, denoted as ${K_s}$ and ${K_v}\in\mathbb{R}^{d_s}$.

\subsection{Knowledge Integration}

To leverage contextual information, we employ a cross-attention mechanism to integrate the inline text with two types of caption information, due to the consistency of the textual vector space.
The feature introducing the textual caption $K_s$ is defined as
$S_{K_s} = \operatorname{Softmax}\left(S {K_s}^T / \sqrt{d_s} \right) S$. 
Similarly, the feature introducing the visual caption ${K_v}$ is obtained and denoted as $S_{K_v}$. 
We then incorporate these features into a knowledge-enhanced representation $S_K = \text{Mean}(S, S_{K_s}, S_{K_v})$, where $S_K \in\mathbb{R}^{d_s}$.
Next, we concatenate $S_K$ with the original image feature $V$ to obtain the final representation of a meme, denoted as $C\in\mathbb{R}^{d_c}$, where $d_c = d_s + d_v$. 
$C$ is then processed by a trainable classifier, which applies a linear transformation followed by a softmax function to produce detection probabilities.
% $P = \operatorname{Softmax}\left( WC + b \right)$, where $W \in\mathbb{R}^{M \times d_c}$ is a trainable parameter matrix $b\in\mathbb{R}^{M}$ is a bias term, $M$ is the number of classes. 

\section{Experiments}
\subsection{Tasks and Baselines}
% To demonstrate the usability of the dataset and provide a benchmark,
We utilize \textsc{ToxiCN MM} as the benchmark for Chinese harmful meme detection.
Specifically, we establish two progressive tasks. (I) \textbf{Harmful Meme Detection}, a binary classification task, detects if a meme is harmful;
(II) \textbf{Harmful Type Identification}, a multi-classification, further identifies its harmful type, including targeted harmful memes, general offense, sexual innuendo, or dispirited culture. 
In addition to MKE, we evaluate the performance of various baselines, including both unimodal and multimodal models.
For unimodal models, we utilize RoBERTa \citep{DBLP:journals/corr/abs-1907-11692}, GPT-3.5, and GPT-4 (text input) as text-only models, while ResNet \citep{DBLP:conf/cvpr/HeZRS16} and ViT \citep{DBLP:conf/iclr/DosovitskiyB0WZ21} serve as image-only models. 
For multimodal models, we employ CLIP \citep{DBLP:conf/icml/RadfordKHRGASAM21}, the fusion of RoBERTa and ViT, which concatenates representations of the text and image for classification, and GPT-4 (text and image input).

\subsection{Implementation}
We adopt precision (\textit{P}), recall (\textit{R}), and macro $F_1$-score ($F_1$) as metrics. 
% In the task of \textit{harmful type identification},
We also report the $F_1$ of harmful memes and each harmful type.
We respectively utilize CLIP and the fusion of RoBERTa and ViT as the backbones of MKE, and we use GPT-4 to generate enhanced captions.
% Both instruction templates and captions are in Chinese.
For conventional PLMs, we fine-tune their parameters and select the best-performing model based on test set outcomes.
For LLMs, we evaluate their performance in a zero-shot scenario, using instruction templates in Chinese.
More details are provided in Appendix \ref{Appendix_exp}.

\subsection{Results and Discussions}
In this section, we present our experimental results and conduct a detailed analysis.
The performance of baselines is evaluated across two tasks, as shown in \tablename~\ref{result}. 
From the results, we can observe that:

\begin{table*}
\renewcommand{\arraystretch}{1.2}
\small
  \centering
  \scalebox{0.85}{
    \begin{tabular}{cl|m{0.7cm}<{\centering}m{0.7cm}<{\centering}m{0.7cm}<{\centering}m{0.7cm}<{\centering}|m{0.7cm}<{\centering}m{0.7cm}<{\centering}m{0.7cm}<{\centering}m{0.7cm}<{\centering}m{0.7cm}<{\centering}m{0.7cm}<{\centering}m{0.7cm}<{\centering}}
    \hline
    \multirow{2}[4]{*}{Modality} & \multirow{2}[4]{*}{Model} & \multicolumn{4}{c|}{Harmful Meme Detection} & \multicolumn{7}{c}{Harmful Type Identification} \\
\cline{3-13}    &       & P     & R     & F1    & $\text{F1}_\text{Harm.}$    & P     & R     & F1    & $\text{F1}_{\text{Tg.}}$ & $\text{F1}_{\text{Off.}}$ & $\text{F1}_{\text{Sex.}}$ & $\text{F1}_{\text{Disp.}}$ \\
    \hline
    \multirow{3}[2]{*}{Text} & GPT3.5 & 69.46 & 66.59 & 67.46 & 53.25 & 36.93 & 32.60 & 32.45 & 27.42 & 29.01 & 10.52 & 14.63 \\
          & GPT4  & 74.52 & 65.59 & 68.01 & 51.78 & 58.29 & 41.81 & 44.86 & 11.43 & 62.22 & 32.26 & 34.15 \\
          & RoBERTa & 75.52 & 77.54 & 76.36 & 66.48 & 53.24 & 60.06 & 55.85 & 48.79 & 71.81 & 42.70 & 29.23 \\
    \hline
    \multirow{2}[2]{*}{Image} & ResNet & 66.61 & 66.92 & 66.76 & 53.76 & 35.66 & 36.23 & 36.46 & 16.67 & 51.28 & 20.93 & 12.28 \\
          & ViT   & 68.97 & 68.61 & 68.78 & 57.24 & 43.38 & 37.86 & 39.10 & 24.17 & 54.57 & 34.32 & 9.01 \\
    \hline
    \multirow{9}[6]{*}{Multimodal} & GPT4  & 74.67 & 68.64 & 70.11 & 55.77 & 58.87 & 41.77 & 43.89 & 23.53 & 32.56 & 55.74 & 23.53 \\
\cline{2-13}          & Fusion & 77.77 & 79.18 & 78.39 & 69.61 & 58.93 & 60.58 & 59.35 & 50.24 & 73.35 & 47.85 & 38.71 \\
          & ~ + $K_s$ & 78.17 & 79.32 & 79.04 & 70.33 & 60.09 & 63.38 & 61.28 & 51.02 & 75.60 & 48.75 & \textbf{43.06} \\
          & ~ + $K_v$ & 77.93 & 79.32 & 78.55 & 69.85 & 59.16 & 60.67 & 59.61 & 51.41 & 75.00 & 48.68 & 36.23 \\
          & ~ + MKE & 77.96 & \underline{80.96} & 79.16 & 70.22 & \textbf{62.41} & 62.08 & \textbf{62.17} & \underline{53.27} & 74.54 & \underline{56.10} & \underline{39.24} \\
\cline{2-13}          & CLIP  & 78.95 & 80.26 & 79.54 & 71.28 & 54.85 & \underline{64.95} & 57.85 & 49.58 & 74.65 & 49.64 & 26.92 \\
          & ~ \ + $K_s$ & 79.23 & 80.71 & 79.89 & 71.72 & 56.24 & \textbf{66.45} & 59.23 & 47.80 & \underline{77.17} & 54.72 & 27.78 \\
          & ~ \ + $K_v$ & \underline{79.39} & \textbf{80.93} & \underline{80.07} & \underline{71.96} & 57.20 & 64.27 & 59.24 & \textbf{53.41} & 76.97 & 52.53 & 25.24 \\
          & ~ \ + MKE & \textbf{79.76} & 80.79 & \textbf{80.23} & \textbf{72.35} & \underline{60.38} & 63.52 & \underline{61.47} & 51.52 & \textbf{77.19} & \textbf{57.14} & 33.33 \\
    \hline
    \end{tabular}%
    }
    % \vspace{-0.05in}
    \caption{Detection performance of baselines. Results show the mean of \textit{P}, \textit{R}, and macro $F_1$, where the \textbf{bold} and \underline{underline} scores respectively represent the optimal and suboptimal values. \textit{Fusion} refers to the fusion of RoBERTa and ViT, and $K_s$ and $K_v$ respectively denote introducing enhanced captions of inline text and image. All results are statistically significant, as determined by a $t$-test ($p < 0.01$).} 
    % 
    % \vspace{-0.1in}
  \label{result}%
\end{table*}%

(1) In contrast to LLMs, conventional fine-tuned pre-trained baselines (i.e., CLIP and the combination of RoBERTa and ViT) achieve better detection performance, indicating their effectiveness in specific tasks.
When considering the modality of input information, RoBERTa, which solely utilizes the inline text of memes, achieves a significantly higher $F_1$ score (average increase of 8.4\%) than vision-based methods such as ResNet and ViT, which solely utilize images.
This result supports the conclusion drawn in \citep{DBLP:conf/www/HeeLC22}, namely that text comprehension plays a more crucial role than image understanding in the detection of harmful memes.

(2) GPT-4 and GPT-3.5 show similar performance in binary \textit{harmful meme detection} when only the inline text is provided, and there is a clear enhancement in the multiclass \textit{harmful type identification} task.
After incorporating the image input, GPT-4 shows the best detection performance for sexual innuendo (\textit{Sex.}) memes, while its performance decreases for general offense (\textit{Off.}) and dispirited culture (\textit{Disp.}).
Referring to Table \ref{harmful_fusion}, we observe that most samples of \textit{Sex.} exhibit toxicity through image-text fusion or harmful images, whereas images of \textit{Off.} and \textit{Disp.} are mostly benign.
This suggests that the toxicity of visual information has a significant impact on the decisions of GPT-4.
We will further explain this in the following case study.

(3) Our MKE demonstrates superior performance, with an average macro-F1 -score improvement of 0.73\% and 3.22\% over the backbone models for both tasks. 
This improvement illustrates the effectiveness of introducing contextual information of meme content for detecting Chinese harmful memes.  
Ablation studies show that both enhanced captions for inline text and images contribute to the detector’s deeper understanding of memes, leading to more precise classifications. 
Additionally, the degree of performance enhancement varies depending on the type of harmful meme. 
For instance, for targeted harmful memes (\textit{Tg.}), where toxicity often relies on the combination of image and text, image captions provide a greater boost (2.07\%). In contrast, for memes expressing dispirited culture (\textit{Disp.}), where the text is typically harmful, inline text captions lead to a larger improvement (2.58\%).

\subsection{Case Study}

To further illustrate the rationales of MKE, we provide several case studies, as shown in Table \ref{case_study}. 
We list enhanced captions of harmful memes and the predictions of other models for reference. 
We also instruct GPT-4 to generate reasons for its detection decisions. 
We do not introduce additional templates to standardize its reasoning to reflect GPT-4's true understanding of memes more accurately.

Exp. (a) is a targeted harmful meme towards Asians.
Through the caption, we observe that GPT-4 understands the meme's meaning solely through the inline text, recognizing the high standard of Asian parents on their children's academic performance. 
This highlights GPT-4's strong contextual understanding. 
After integrating this information, compared to the backbone (Fusion), our MKE model makes the correct decision, illustrating that incorporating contextual information of meme content enhances the model's understanding of memes.

For more insight into the challenges of detecting Chinese harmful memes, we manually inspected the samples misclassified by most baselines. Two main types of errors are summarized. 

\textbf{Type I error}: Benign information contained in harmful memes can influence the judgment of models, resulting in incorrect detection.
In Exp. (b), when presented solely with the inline text, GPT-4 accurately interprets the meme's meaning, comparing "\textit{I}" with a ”\textit{little mouse}” to convey a dispirited culture.
However, upon introducing the image, GPT-4 mistakenly interprets the mouse as being "\textit{gently stroked}" and incorrectly categorizes the meme as harmless.
This suggests that the model may overlook the potential toxicity of memes due to the seemingly benign nature of a certain modal.
In contrast, MKE integrates original sample and caption information to make the correct judgment.
% This suggests that the model may hallucinate during multimodal reasoning, leading to an oversight of potential toxicity in the meme.

\textbf{Type II error}: Harmful memes containing unique cultural backgrounds are easily missed by models.
Among them, the most challenging samples are the memes whose inline text contains specific expressions in Chinese, such as homophones and metaphors.
Take Exp. (c) for example, where the term "\textit{dog-banana}" is a homonym of "\textit{bark}" in Chinese.
Therefore, this meme is essentially a harmful meme containing general offense, implicitly expressing dissatisfaction with another person.
Due to the lack of related knowledge, current models fail to comprehend the underlying semantics of these memes, making them difficult to detect successfully.

These case studies further illustrate that Chinese harmful memes detection is a complex multimodal semantic understanding task, which is challenging for existing models.
The error analysis shows that fully integrating image and text information and introducing more comprehensive knowledge about Chinese culture are both crucial for effectively detecting harmful memes.

\begin{table*}
  \centering
  \scalebox{0.87}{
  \begin{tabular}{m{1.5cm}<{\centering}|m{4.25cm}|m{4.25cm}|m{4.25cm}<{\centering}}
    \hline
    \rule{0pt}{8pt}
    \small Harmful Meme & \centering\includegraphics[width=4.25cm]{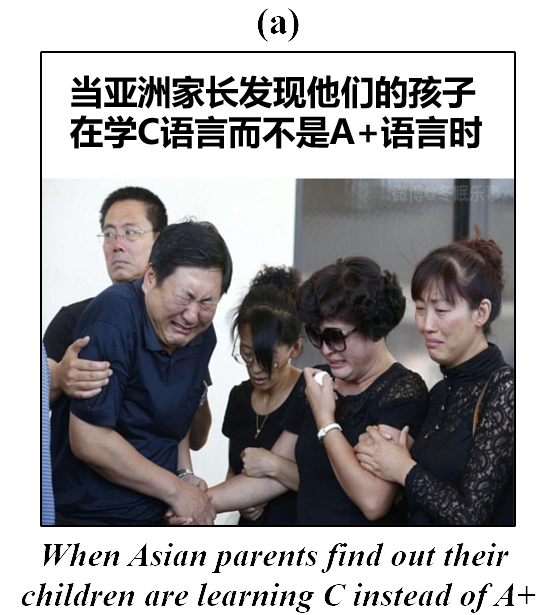} & \centering\includegraphics[width=3.7cm]{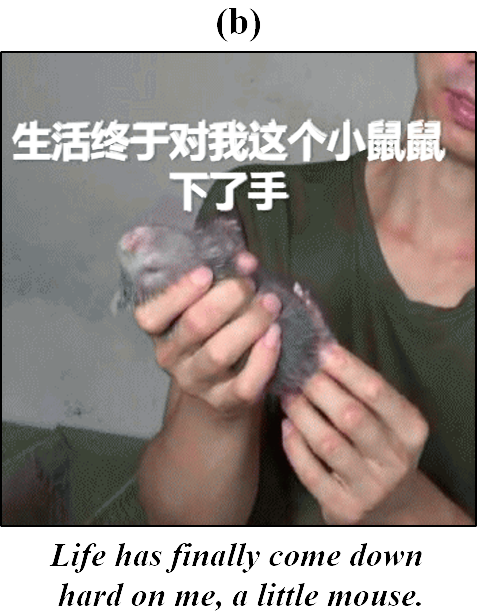} & \includegraphics[width=3.8cm]{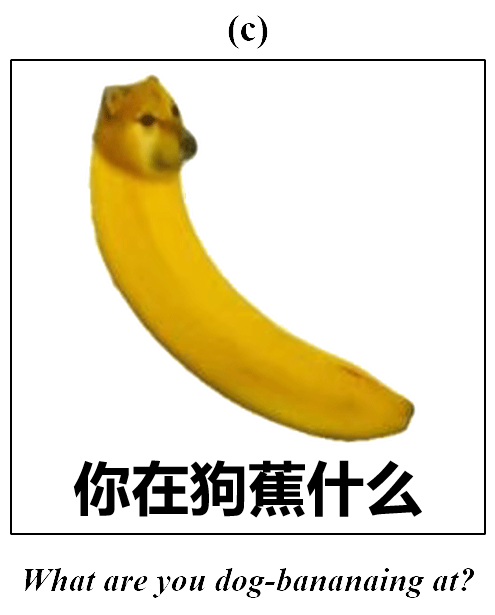} \\
    \hline
    \small Category & \scriptsize \centering{Targeted Harmful, Text-Image Fusion} & \scriptsize \centering{Dispirited Culture, Harmful Text} & \scriptsize General Offensive, Harmful Text\\
    \hline
    \small Text Caption &  \scriptsize Asian parents demand \textcolor{LimeGreen}{high academic performance}, expecting their children to get excellence (A+) and not average grades (C). & \scriptsize It means that life is stressful or difficult. It implies that one is as small or unimportant as a mouse, \textcolor{LimeGreen}{involving dispirited culture}.
 & \multicolumn{1}{m{4.25cm}}{\scriptsize "\textit{狗蕉}" may be a slang term, and the text meaning may be jokingly \textcolor{Red}{describing some combination of dog and banana.}}\\
    \hline
    \small Image Caption & \scriptsize Four \textcolor{LimeGreen}{grieving Asian adults} were \textcolor{LimeGreen}{crying} and holding each other up. & \scriptsize One man \textcolor{Red}{gently} cradled a small hamster-like animal in his hand. & \scriptsize The image shows a yellow banana with one end replaced with a pattern of a dog's head. \\
    \hline
    \small Explanation & \scriptsize This meme can be considered a humorous reference to \textcolor{LimeGreen}{cultural stereotypes}, exaggerating Asian parents' concern for their children's academic performance. & \scriptsize The meaning of this meme is to make a self-joke about the hardships of life. It is \textcolor{Red}{just a humorous way} to express someone's helpless sense of life \textcolor{Red}{without toxicity}. & \multicolumn{1}{m{4.25cm}}{\scriptsize This meme is \textcolor{Red}{just a kind of humor}. By combining the dog and banana together with words, it produces a humorous effect that subverts expectations.} \\
    \hline
    \small Prediction & \scriptsize GPT-4 (only text): \ding{51}, GPT-4: \ding{51}, CLIP: \ding{51}, Fusion: \ding{55}, MKE (Fusion): \ding{51}. & \scriptsize GPT-4 (only text): \ding{51}, GPT-4: \ding{55}, CLIP: \ding{55}, Fusion: \ding{55}, MKE (Fusion): \ding{51}. & \multicolumn{1}{m{4.25cm}}{\scriptsize GPT-4 (only text): \ding{55}, GPT-4: \ding{55}, CLIP: \ding{55}, Fusion: \ding{55}, MKE (Fusion): \ding{55}.} \\
    \hline
  \end{tabular}
  }
  % \vspace{-0.05in}
  \caption{Illustration of case study. Highlighted in \textcolor{LimeGreen}{green} and \textcolor{Red}{red} within the decision reasons are the accurate implications of toxicity and misinformation. \ding{51} and \ding{55} represent the success and failure of model prediction. In the implementation, all descriptions are provided in Chinese.}
  % \vspace{-0.2in}
  \label{case_study}
\end{table*}

\section{Conclusions and Future Work}
In this paper, we focus on the comprehensive detection of Chinese harmful memes.
We present the first Chinese harmful meme dataset \textsc{ToxiCN MM}.
It has 12k samples including not only targeted harmful memes but also those only exhibiting potential toxicity without specific targets, adapting to the Chinese online environment.
In addition to binary labels, \textsc{ToxiCN MM} provides harmful types and modality combination categories of memes.
To improve the understanding of Chinese harmful memes, we present a Multimodal Knowledge Enhancement (MKE) detector, introducing the contextual information of inline text and images.
In the experimental phase, we evaluate multiple baseline models for their performance in detecting Chinese harmful memes.
Our case study suggests that integrating multimodal information and comprehensive background knowledge is crucial for effective detection.

In future work, we aim to design more effective methods for Chinese harmful memes detection.
Meanwhile, we notice that the accuracy of LLMs in detecting Chinese harmful memes is still limited. 
Considering the potential harm that these memes may cause, this task can be used to evaluate the safety of LLMs. 
We will employ prompt engineering and instruction fine-tuning methods to explore and enhance the detection performance of LLMs.
Additionally, we will continuously evaluate state-of-the-art models to ensure the effectiveness of \textsc{ToxiCN MM}.
We expect our dataset, benchmark, and insights will assist researchers in related fields.

\section{Limitations}

In this study, we focus on several most common harmful types of memes on the Chinese online environment.
Due to the filtering mechanism, some harmful memes, such as those containing \textit{fake news}, are extremely scarce on Chinese platforms. 
As a result, our \textsc{ToxiCN MM} does not encompass all harmful types. 
The techniques we used to boost the percentage of harmful content during the dataset construction process may introduce problematic bias.
In future work, we plan to broaden the scope and increase the number of meme crawls, focusing on more Chinese platforms to mitigate sampling bias.
While we have implemented several measures to mitigate annotation bias, we acknowledge that our dataset may still contain mislabeled data due to the subjective understanding of annotators for Chinese harmful memes.
Furthermore, our current study primarily focuses on predicting whether a given meme is harmful.
We will further evaluate the ability of baselines to generate explanations for Chinese harmful memes with quantitative experiments.

\section{Ethics Statement}
Our study aims to facilitate the comprehensive detection of Chinese harmful memes and raise researchers' attention to non-English memes.
The social psychological community has recognized the harms of the harmful types we selected in the dataset.
We acknowledge the risk of malicious actors attempting to reverse-engineer memes. We strongly discourage and denounce such practices, emphasizing the necessity of human moderation to prevent them. All resources are intended solely for scientific research and are prohibited from commercial use. We believe the benefits of our proposed resources outweigh the associated risks.
We strictly follow the data use agreements of each public online social platform.
% It is important to note that all data has been anonymized and does not include any personal information.
The opinions and findings contained in the samples of our presented dataset should not be interpreted as representing the views expressed or implied by the authors. 

To mitigate the potential psychological impact on annotators evaluating harmful content, we implement the following protective measures: 1) obtain explicit consent regarding exposure to potentially abusive content, 2) limit weekly evaluations to manage exposure and ensure reasonable daily workloads, and 3) recommend discontinuing reviews if they experience distress. Additionally, we conduct regular well-being checks to monitor their mental health.

% \section*{Ethics Statement}

% We adhere strictly to the data use agreements of each Chinese social platform and double-check to ensure that no data involving user privacy is included. 
% The opinions and findings contained in the samples of our presented dataset should not be construed as representing the views expressed or implied by the authors. 
% We aim for the benefits of our proposed resources to outweigh their risks. 
% All resources are intended solely for scientific research purposes.

\section*{Acknowledgment}
This research is supported by the Natural Science Foundation of China (No. 62376051, 62076046, 62076051), the Liaoning Province Applied Basic Research Program (No. 2022JH2/101300270), the Liaoning Provincial Natural Science Foundation Joint Fund Program(2023-MSBA-003), and the Fundamental Research Funds for the Central Universities (DUT24MS003).
We would like to thank all reviewers for their constructive comments.

\bibliography{anthology, basic_for_TLD, multimodal_hate, multimodal, custom}

\begin{thebibliography}{44}
\expandafter\ifx\csname natexlab\endcsname\relax\def\natexlab#1{#1}\fi

\bibitem[{Aggarwal et~al.(2023)Aggarwal, Chawla, Das, Saha, Mathew, Zesch, and Mukherjee}]{DBLP:conf/www/AggarwalCDSMZ023}
Piush Aggarwal, Pranit Chawla, Mithun Das, Punyajoy Saha, Binny Mathew, Torsten Zesch, and Animesh Mukherjee. 2023.
\newblock \href {https://doi.org/10.1145/3543507.3583356} {Hateproof: Are hateful meme detection systems really robust?}
\newblock In \emph{Proceedings of the {ACM} Web Conference 2023, {WWW} 2023, Austin, TX, USA, 30 April 2023 - 4 May 2023}, pages 3734--3743. {ACM}.

\bibitem[{Bai et~al.(2023)Bai, Bai, Yang, Wang, Tan, Wang, Lin, Zhou, and Zhou}]{Qwen-VL}
Jinze Bai, Shuai Bai, Shusheng Yang, Shijie Wang, Sinan Tan, Peng Wang, Junyang Lin, Chang Zhou, and Jingren Zhou. 2023.
\newblock Qwen-vl: A versatile vision-language model for understanding, localization, text reading, and beyond.
\newblock \emph{arXiv preprint arXiv:2308.12966}.

\bibitem[{Bell(1997)}]{bell1997innuendo}
David~M Bell. 1997.
\newblock Innuendo.
\newblock \emph{Journal of Pragmatics}, 27(1):35--59.

\bibitem[{Blaier et~al.(2021)Blaier, Malkiel, and Wolf}]{DBLP:conf/emnlp/BlaierMW21}
Efrat Blaier, Itzik Malkiel, and Lior Wolf. 2021.
\newblock \href {https://doi.org/10.18653/v1/2021.emnlp-main.738} {Caption enriched samples for improving hateful memes detection}.
\newblock In \emph{Proceedings of the 2021 Conference on Empirical Methods in Natural Language Processing, {EMNLP} 2021, Virtual Event / Punta Cana, Dominican Republic, 7-11 November, 2021}, pages 9350--9358. Association for Computational Linguistics.

\bibitem[{Cao et~al.(2022)Cao, Lee, Chong, and Jiang}]{DBLP:conf/emnlp/CaoLC022}
Rui Cao, Roy~Ka{-}Wei Lee, Wen{-}Haw Chong, and Jing Jiang. 2022.
\newblock \href {https://doi.org/10.18653/v1/2022.emnlp-main.22} {Prompting for multimodal hateful meme classification}.
\newblock In \emph{Proceedings of the 2022 Conference on Empirical Methods in Natural Language Processing, {EMNLP} 2022, Abu Dhabi, United Arab Emirates, December 7-11, 2022}, pages 321--332. Association for Computational Linguistics.

\bibitem[{Dixon et~al.(2018)Dixon, Li, Sorensen, Thain, and Vasserman}]{DBLP:conf/aies/DixonLSTV18}
Lucas Dixon, John Li, Jeffrey Sorensen, Nithum Thain, and Lucy Vasserman. 2018.
\newblock \href {https://doi.org/10.1145/3278721.3278729} {Measuring and mitigating unintended bias in text classification}.
\newblock In \emph{Proceedings of the 2018 {AAAI/ACM} Conference on AI, Ethics, and Society, {AIES} 2018, New Orleans, LA, USA, February 02-03, 2018}, pages 67--73. {ACM}.

\bibitem[{Dong et~al.(2017)Dong, Chang, and Sun}]{QSNY201703001}
Ziyang Dong, Jinfeng Chang, and Jian Sun. 2017.
\newblock \href {https://doi.org/10.16399/j.cnki.qsnyj.2017.03.001} {A study on "dispirited culture" of online youth from the perspective of social psychology}.
\newblock \emph{Youth and Adolescent Studies}, (3-7+31).

\bibitem[{Dosovitskiy et~al.(2021)Dosovitskiy, Beyer, Kolesnikov, Weissenborn, Zhai, Unterthiner, Dehghani, Minderer, Heigold, Gelly, Uszkoreit, and Houlsby}]{DBLP:conf/iclr/DosovitskiyB0WZ21}
Alexey Dosovitskiy, Lucas Beyer, Alexander Kolesnikov, Dirk Weissenborn, Xiaohua Zhai, Thomas Unterthiner, Mostafa Dehghani, Matthias Minderer, Georg Heigold, Sylvain Gelly, Jakob Uszkoreit, and Neil Houlsby. 2021.
\newblock \href {https://openreview.net/forum?id=YicbFdNTTy} {An image is worth 16x16 words: Transformers for image recognition at scale}.
\newblock In \emph{9th International Conference on Learning Representations, {ICLR} 2021, Virtual Event, Austria, May 3-7, 2021}. OpenReview.net.

\bibitem[{Du et~al.(2022)Du, Qian, Liu, Ding, Qiu, Yang, and Tang}]{du2022glm}
Zhengxiao Du, Yujie Qian, Xiao Liu, Ming Ding, Jiezhong Qiu, Zhilin Yang, and Jie Tang. 2022.
\newblock Glm: General language model pretraining with autoregressive blank infilling.
\newblock In \emph{Proceedings of the 60th Annual Meeting of the Association for Computational Linguistics (Volume 1: Long Papers)}, pages 320--335.

\bibitem[{Fersini et~al.(2022)Fersini, Gasparini, Rizzi, Saibene, Chulvi, Rosso, Lees, and Sorensen}]{DBLP:conf/semeval/FersiniGRSCRLS22}
Elisabetta Fersini, Francesca Gasparini, Giulia Rizzi, Aurora Saibene, Berta Chulvi, Paolo Rosso, Alyssa Lees, and Jeffrey Sorensen. 2022.
\newblock \href {https://doi.org/10.18653/v1/2022.semeval-1.74} {Semeval-2022 task 5: Multimedia automatic misogyny identification}.
\newblock In \emph{Proceedings of the 16th International Workshop on Semantic Evaluation, SemEval@NAACL 2022, Seattle, Washington, United States, July 14-15, 2022}, pages 533--549. Association for Computational Linguistics.

\bibitem[{Fortuna et~al.(2022)Fortuna, Dom{\'{\i}}nguez, Wanner, and Talat}]{DBLP:conf/emnlp/FortunaDWT22}
Paula Fortuna, M{\'{o}}nica Dom{\'{\i}}nguez, Leo Wanner, and Zeerak Talat. 2022.
\newblock \href {https://aclanthology.org/2022.emnlp-main.809} {Directions for {NLP} practices applied to online hate speech detection}.
\newblock In \emph{Proceedings of the 2022 Conference on Empirical Methods in Natural Language Processing, {EMNLP} 2022, Abu Dhabi, United Arab Emirates, December 7-11, 2022}, pages 11794--11805. Association for Computational Linguistics.

\bibitem[{Gomez et~al.(2020)Gomez, Gibert, G{\'{o}}mez, and Karatzas}]{DBLP:conf/wacv/GomezGGK20}
Raul Gomez, Jaume Gibert, Llu{\'{\i}}s G{\'{o}}mez, and Dimosthenis Karatzas. 2020.
\newblock \href {https://doi.org/10.1109/WACV45572.2020.9093414} {Exploring hate speech detection in multimodal publications}.
\newblock In \emph{{IEEE} Winter Conference on Applications of Computer Vision, {WACV} 2020, Snowmass Village, CO, USA, March 1-5, 2020}, pages 1459--1467. {IEEE}.

\bibitem[{He et~al.(2016)He, Zhang, Ren, and Sun}]{DBLP:conf/cvpr/HeZRS16}
Kaiming He, Xiangyu Zhang, Shaoqing Ren, and Jian Sun. 2016.
\newblock \href {https://doi.org/10.1109/CVPR.2016.90} {Deep residual learning for image recognition}.
\newblock In \emph{2016 {IEEE} Conference on Computer Vision and Pattern Recognition, {CVPR} 2016, Las Vegas, NV, USA, June 27-30, 2016}, pages 770--778. {IEEE} Computer Society.

\bibitem[{Hee et~al.(2022)Hee, Lee, and Chong}]{DBLP:conf/www/HeeLC22}
Ming~Shan Hee, Roy~Ka{-}Wei Lee, and Wen{-}Haw Chong. 2022.
\newblock \href {https://doi.org/10.1145/3485447.3512260} {On explaining multimodal hateful meme detection models}.
\newblock In \emph{{WWW} '22: The {ACM} Web Conference 2022, Virtual Event, Lyon, France, April 25 - 29, 2022}, pages 3651--3655. {ACM}.

\bibitem[{Hossain et~al.(2022)Hossain, Sharif, and Hoque}]{DBLP:conf/ijcnlp/HossainSH22}
Eftekhar Hossain, Omar Sharif, and Mohammed~Moshiul Hoque. 2022.
\newblock \href {https://aclanthology.org/2022.aacl-srw.5} {{MUTE:} {A} multimodal dataset for detecting hateful memes}.
\newblock In \emph{Proceedings of the 2nd Conference of the Asia-Pacific Chapter of the Association for Computational Linguistics and the 12th International Joint Conference on Natural Language Processing, {AACL/IJCNLP} 2022 - Student Research Workshop, Online, November 20, 2022}, pages 32--39. Association for Computational Linguistics.

\bibitem[{Hu et~al.(2023)Hu, Yao, Wang, Wang, Pan, Chen, Yu, Wu, Zhao, Zhang, Han, Lin, Xue, Li, Liu, and Sun}]{DBLP:journals/corr/abs-2308-12038}
Jinyi Hu, Yuan Yao, Chongyi Wang, Shan Wang, Yinxu Pan, Qianyu Chen, Tianyu Yu, Hanghao Wu, Yue Zhao, Haoye Zhang, Xu~Han, Yankai Lin, Jiao Xue, Dahai Li, Zhiyuan Liu, and Maosong Sun. 2023.
\newblock \href {https://doi.org/10.48550/ARXIV.2308.12038} {Large multilingual models pivot zero-shot multimodal learning across languages}.
\newblock \emph{CoRR}, abs/2308.12038.

\bibitem[{Ji et~al.(2023)Ji, Ren, and Naseem}]{DBLP:conf/www/JiRN23}
Junhui Ji, Wei Ren, and Usman Naseem. 2023.
\newblock \href {https://doi.org/10.1145/3543507.3587427} {Identifying creative harmful memes via prompt based approach}.
\newblock In \emph{Proceedings of the {ACM} Web Conference 2023, {WWW} 2023, Austin, TX, USA, 30 April 2023 - 4 May 2023}, pages 3868--3872. {ACM}.

\bibitem[{Kiela et~al.(2020)Kiela, Firooz, Mohan, Goswami, Singh, Ringshia, and Testuggine}]{DBLP:conf/nips/KielaFMGSRT20}
Douwe Kiela, Hamed Firooz, Aravind Mohan, Vedanuj Goswami, Amanpreet Singh, Pratik Ringshia, and Davide Testuggine. 2020.
\newblock \href {https://proceedings.neurips.cc/paper/2020/hash/1b84c4cee2b8b3d823b30e2d604b1878-Abstract.html} {The hateful memes challenge: Detecting hate speech in multimodal memes}.
\newblock In \emph{Advances in Neural Information Processing Systems 33: Annual Conference on Neural Information Processing Systems 2020, NeurIPS 2020, December 6-12, 2020, virtual}.

\bibitem[{Koukounas and Letch(2001)}]{koukounas2001psychological}
Eric Koukounas and Nicole~M Letch. 2001.
\newblock Psychological correlates of perception of sexual intent in women.
\newblock \emph{The Journal of social psychology}, 141(4):443--456.

\bibitem[{Koutlis et~al.(2023)Koutlis, Schinas, and Papadopoulos}]{DBLP:conf/mir/KoutlisSP23}
Christos Koutlis, Manos Schinas, and Symeon Papadopoulos. 2023.
\newblock \href {https://doi.org/10.1145/3591106.3592254} {Memefier: Dual-stage modality fusion for image meme classification}.
\newblock In \emph{Proceedings of the 2023 {ACM} International Conference on Multimedia Retrieval, {ICMR} 2023, Thessaloniki, Greece, June 12-15, 2023}, pages 586--591. {ACM}.

\bibitem[{Li et~al.(2022)Li, Lin, Yang, Xu, and Zhang}]{DBLP:conf/nlpcc/LiLYXZ22}
Zefeng Li, Hongfei Lin, Liang Yang, Bo~Xu, and Shaowu Zhang. 2022.
\newblock \href {https://doi.org/10.1007/978-3-031-17120-8\_41} {Memeplate: {A} chinese multimodal dataset for humor understanding in meme templates}.
\newblock In \emph{Natural Language Processing and Chinese Computing - 11th {CCF} International Conference, {NLPCC} 2022, Guilin, China, September 24-25, 2022, Proceedings, Part {I}}, volume 13551 of \emph{Lecture Notes in Computer Science}, pages 527--538. Springer.

\bibitem[{Lin and Zhang(2019)}]{lin2019}
Aijun Lin and Bo~Zhang. 2019.
\newblock Emojis as discourse: Symbolic consumption and sociological reflection on internet emojis.
\newblock \emph{Modern Communication (Journal of Communication University of China)}, 41(35-40).

\bibitem[{Liu and Xu(2016)}]{liu2016}
Min Liu and Shuai Xu. 2016.
\newblock An instant 'meme battle': Communication and identification in the spread of emoji packs".
\newblock \emph{Journal of News Research}, 7(339).

\bibitem[{Liu et~al.(2019)Liu, Ott, Goyal, Du, Joshi, Chen, Levy, Lewis, Zettlemoyer, and Stoyanov}]{DBLP:journals/corr/abs-1907-11692}
Yinhan Liu, Myle Ott, Naman Goyal, Jingfei Du, Mandar Joshi, Danqi Chen, Omer Levy, Mike Lewis, Luke Zettlemoyer, and Veselin Stoyanov. 2019.
\newblock \href {http://arxiv.org/abs/1907.11692} {Roberta: {A} robustly optimized {BERT} pretraining approach}.
\newblock \emph{CoRR}, abs/1907.11692.

\bibitem[{Lu et~al.(2023)Lu, Xu, Zhang, Min, Yang, and Lin}]{DBLP:conf/acl/LuXZMYL23}
Junyu Lu, Bo~Xu, Xiaokun Zhang, Changrong Min, Liang Yang, and Hongfei Lin. 2023.
\newblock \href {https://doi.org/10.18653/v1/2023.acl-long.898} {Facilitating fine-grained detection of chinese toxic language: Hierarchical taxonomy, resources, and benchmarks}.
\newblock In \emph{Proceedings of the 61st Annual Meeting of the Association for Computational Linguistics (Volume 1: Long Papers), {ACL} 2023, Toronto, Canada, July 9-14, 2023}, pages 16235--16250. Association for Computational Linguistics.

\bibitem[{Miao and Xu(2022)}]{LLDK202202018}
Cunlong Miao and Maohua Xu. 2022.
\newblock The representation and guidance path of youth "dispirited culture".
\newblock \emph{Journal of Socialist Theory Guide}, (123-128).

\bibitem[{Otsri(2020)}]{otsri2020non}
Magi Otsri. 2020.
\newblock Non-sexist sexual humor as quid pro quo sexual harassment.
\newblock \emph{Sexuality \& Culture}, 24(1):94--112.

\bibitem[{Peng(2019)}]{peng2019}
Lan Peng. 2019.
\newblock \href {https://doi.org/10.15896/j.xjtuskxb.201901012} {Emotion icon: Password, label and mask}.
\newblock \emph{Journal of Xi’an Jiaotong University (Social Sciences)}, 39(104-110+153).

\bibitem[{Pramanick et~al.(2021{\natexlab{a}})Pramanick, Dimitrov, Mukherjee, Sharma, Akhtar, Nakov, and Chakraborty}]{DBLP:conf/acl/PramanickDMSANC21}
Shraman Pramanick, Dimitar Dimitrov, Rituparna Mukherjee, Shivam Sharma, Md.~Shad Akhtar, Preslav Nakov, and Tanmoy Chakraborty. 2021{\natexlab{a}}.
\newblock \href {https://doi.org/10.18653/v1/2021.findings-acl.246} {Detecting harmful memes and their targets}.
\newblock In \emph{Findings of the Association for Computational Linguistics: {ACL/IJCNLP} 2021, Online Event, August 1-6, 2021}, volume {ACL/IJCNLP} 2021 of \emph{Findings of {ACL}}, pages 2783--2796. Association for Computational Linguistics.

\bibitem[{Pramanick et~al.(2021{\natexlab{b}})Pramanick, Sharma, Dimitrov, Akhtar, Nakov, and Chakraborty}]{DBLP:conf/emnlp/PramanickSDAN021}
Shraman Pramanick, Shivam Sharma, Dimitar Dimitrov, Md.~Shad Akhtar, Preslav Nakov, and Tanmoy Chakraborty. 2021{\natexlab{b}}.
\newblock \href {https://doi.org/10.18653/v1/2021.findings-emnlp.379} {{MOMENTA:} {A} multimodal framework for detecting harmful memes and their targets}.
\newblock In \emph{Findings of the Association for Computational Linguistics: {EMNLP} 2021, Virtual Event / Punta Cana, Dominican Republic, 16-20 November, 2021}, pages 4439--4455. Association for Computational Linguistics.

\bibitem[{Radford et~al.(2021)Radford, Kim, Hallacy, Ramesh, Goh, Agarwal, Sastry, Askell, Mishkin, Clark, Krueger, and Sutskever}]{DBLP:conf/icml/RadfordKHRGASAM21}
Alec Radford, Jong~Wook Kim, Chris Hallacy, Aditya Ramesh, Gabriel Goh, Sandhini Agarwal, Girish Sastry, Amanda Askell, Pamela Mishkin, Jack Clark, Gretchen Krueger, and Ilya Sutskever. 2021.
\newblock \href {http://proceedings.mlr.press/v139/radford21a.html} {Learning transferable visual models from natural language supervision}.
\newblock In \emph{Proceedings of the 38th International Conference on Machine Learning, {ICML} 2021, 18-24 July 2021, Virtual Event}, volume 139 of \emph{Proceedings of Machine Learning Research}, pages 8748--8763. {PMLR}.

\bibitem[{Ross et~al.(2016)Ross, Rist, Carbonell, Cabrera, Kurowsky, and Wojatzki}]{ross2017measuring}
Bjorn Ross, Michael Rist, Guillermo Carbonell, Benjamin Cabrera, Nils Kurowsky, and Michael Wojatzki. 2016.
\newblock Measuring the reliability of hate speech annotations: The case of the european refugee crisis.
\newblock In \emph{3rd Workshop on Natural Language Processing for Computer-Mediated Communication/Social Media}, pages 6--9. Ruhr-Universitat Bochum.

\bibitem[{Sharma et~al.(2022{\natexlab{a}})Sharma, Akhtar, Nakov, and Chakraborty}]{DBLP:conf/naacl/SharmaAN022}
Shivam Sharma, Md.~Shad Akhtar, Preslav Nakov, and Tanmoy Chakraborty. 2022{\natexlab{a}}.
\newblock \href {https://doi.org/10.18653/v1/2022.findings-naacl.118} {{DISARM:} detecting the victims targeted by harmful memes}.
\newblock In \emph{Findings of the Association for Computational Linguistics: {NAACL} 2022, Seattle, WA, United States, July 10-15, 2022}, pages 1572--1588. Association for Computational Linguistics.

\bibitem[{Sharma et~al.(2022{\natexlab{b}})Sharma, Alam, Akhtar, Dimitrov, Martino, Firooz, Halevy, Silvestri, Nakov, and Chakraborty}]{DBLP:conf/ijcai/SharmaAADMFHSN022}
Shivam Sharma, Firoj Alam, Md.~Shad Akhtar, Dimitar Dimitrov, Giovanni Da~San Martino, Hamed Firooz, Alon~Y. Halevy, Fabrizio Silvestri, Preslav Nakov, and Tanmoy Chakraborty. 2022{\natexlab{b}}.
\newblock \href {https://doi.org/10.24963/ijcai.2022/781} {Detecting and understanding harmful memes: {A} survey}.
\newblock In \emph{Proceedings of the Thirty-First International Joint Conference on Artificial Intelligence, {IJCAI} 2022, Vienna, Austria, 23-29 July 2022}, pages 5597--5606. ijcai.org.

\bibitem[{Suryawanshi et~al.(2020)Suryawanshi, Chakravarthi, Verma, Arcan, McCrae, and Buitelaar}]{suryawanshi-etal-2020-dataset}
Shardul Suryawanshi, Bharathi~Raja Chakravarthi, Pranav Verma, Mihael Arcan, John~Philip McCrae, and Paul Buitelaar. 2020.
\newblock \href {https://aclanthology.org/2020.wildre-1.2} {A dataset for troll classification of {T}amil{M}emes}.
\newblock In \emph{Proceedings of the WILDRE5{--} 5th Workshop on Indian Language Data: Resources and Evaluation}, pages 7--13, Marseille, France. European Language Resources Association (ELRA).

\bibitem[{Thornhill and Thornhill(1983)}]{thornhill1983human}
Randy Thornhill and Nancy~Wilmsen Thornhill. 1983.
\newblock Human rape: An evolutionary analysis.
\newblock \emph{Ethology and sociobiology}, 4(3):137--173.

\bibitem[{Wang et~al.(2023)Wang, Li, Lu, Yang, Xia, and Lin}]{DBLP:conf/nlpcc/WangLLYXL23}
Hongbo Wang, Mingda Li, Junyu Lu, Liang Yang, Hebin Xia, and Hongfei Lin. 2023.
\newblock \href {https://doi.org/10.1007/978-3-031-44696-2\_50} {{CCPC:} {A} hierarchical chinese corpus for patronizing and condescending language detection}.
\newblock In \emph{Natural Language Processing and Chinese Computing - 12th National {CCF} Conference, {NLPCC} 2023, Foshan, China, October 12-15, 2023, Proceedings, Part {II}}, volume 14303 of \emph{Lecture Notes in Computer Science}, pages 640--652. Springer.

\bibitem[{Waseem and Hovy(2016)}]{DBLP:conf/naacl/WaseemH16}
Zeerak Waseem and Dirk Hovy. 2016.
\newblock \href {https://doi.org/10.18653/v1/n16-2013} {Hateful symbols or hateful people? predictive features for hate speech detection on twitter}.
\newblock In \emph{Proceedings of the Student Research Workshop, SRW@HLT-NAACL 2016, The 2016 Conference of the North American Chapter of the Association for Computational Linguistics: Human Language Technologies, San Diego California, USA, June 12-17, 2016}, pages 88--93. The Association for Computational Linguistics.

\bibitem[{Xu et~al.(2022)Xu, Li, Zheng, Naseriparsa, Zhao, Lin, and Xia}]{DBLP:conf/sigir/XuLZNZL022}
Bo~Xu, Tingting Li, Junzhe Zheng, Mehdi Naseriparsa, Zhehuan Zhao, Hongfei Lin, and Feng Xia. 2022.
\newblock \href {https://doi.org/10.1145/3477495.3532019} {Met-meme: {A} multimodal meme dataset rich in metaphors}.
\newblock In \emph{{SIGIR} '22: The 45th International {ACM} {SIGIR} Conference on Research and Development in Information Retrieval, Madrid, Spain, July 11 - 15, 2022}, pages 2887--2899. {ACM}.

\bibitem[{Yang(2018)}]{yang2018}
Jianhua Yang. 2018.
\newblock The guidance and norms of the meme culture.
\newblock \emph{People's Tribune}, (140-141).

\bibitem[{Zeinert et~al.(2021)Zeinert, Inie, and Derczynski}]{DBLP:conf/acl/ZeinertID20}
Philine Zeinert, Nanna Inie, and Leon Derczynski. 2021.
\newblock \href {https://doi.org/10.18653/v1/2021.acl-long.247} {Annotating online misogyny}.
\newblock In \emph{Proceedings of the 59th Annual Meeting of the Association for Computational Linguistics and the 11th International Joint Conference on Natural Language Processing, {ACL/IJCNLP} 2021, (Volume 1: Long Papers), Virtual Event, August 1-6, 2021}, pages 3181--3197. Association for Computational Linguistics.

\bibitem[{Zhang and Zhao(2021)}]{zhang2021}
Shengnan Zhang and Linyun Zhao. 2021.
\newblock \href {https://doi.org/10.15997/j.cnki.qnjz.2021.20.005} {The communication mechanism and rational reflection of internet "dispirited culture"}.
\newblock \emph{Youth Journalist}, (40-42).

\bibitem[{Zhao et~al.(2023)Zhao, Zhou, Li, Tang, Wang, Hou, Min, Zhang, Zhang, Dong et~al.}]{zhao2023survey}
Wayne~Xin Zhao, Kun Zhou, Junyi Li, Tianyi Tang, Xiaolei Wang, Yupeng Hou, Yingqian Min, Beichen Zhang, Junjie Zhang, Zican Dong, et~al. 2023.
\newblock A survey of large language models.
\newblock \emph{arXiv preprint arXiv:2303.18223}.

\bibitem[{Zheng(2016)}]{zheng2016}
Manning Zheng. 2016.
\newblock \href {https://doi.org/10.13786/j.cnki.cn14-1066/g2.2016.08.008} {Research on the popularity of network expression meme and its turning of discourse space}.
\newblock \emph{Editorial Friend}, (42-46).

\end{thebibliography}
\bibliographystyle{acl_natbib}

%%%%%%%%%%%%%%%%%%%%%%%%%%%%%%%%%%%%%%%%%%%%%%%%%%%%%%%%%%%%
\newpage
\section*{Checklist}

%%% BEGIN INSTRUCTIONS %%%
% The checklist follows the references.  Please
% read the checklist guidelines carefully for information on how to answer these
% questions.  For each question, change the default \answerTODO{} to \answerYes{},
% \answerNo{}, or \answerNA{}.  You are strongly encouraged to include a {\bf
% justification to your answer}, either by referencing the appropriate section of
% your paper or providing a brief inline description.  For example:
% \begin{itemize}
%   \item Did you include the license to the code and datasets? \answerYes{See Section~\ref{gen_inst}.}
%   \item Did you include the license to the code and datasets? \answerNo{The code and the data are proprietary.}
%   \item Did you include the license to the code and datasets? \answerNA{}
% \end{itemize}
% Please do not modify the questions and only use the provided macros for your
% answers.  Note that the Checklist section does not count towards the page
% limit.  In your paper, please delete this instructions block and only keep the
% Checklist section heading above along with the questions/answers below.
%%% END INSTRUCTIONS %%%

\begin{enumerate}

\item For all authors...
\begin{enumerate}
  \item Do the main claims made in the abstract and introduction accurately reflect the paper's contributions and scope?
    \answerYes{}
  \item Did you describe the limitations of your work?
    \answerYes{See Appendix A.}
  \item Did you discuss any potential negative societal impacts of your work?
    \answerYes{See Appendix B.}
  \item Have you read the ethics review guidelines and ensured that your paper conforms to them?
    \answerYes{}
\end{enumerate}

\item If you are including theoretical results...
\begin{enumerate}
  \item Did you state the full set of assumptions of all theoretical results?
    \answerNA{}
	\item Did you include complete proofs of all theoretical results?
    \answerNA{}
\end{enumerate}

\item If you ran experiments (e.g. for benchmarks)...
\begin{enumerate}
  \item Did you include the code, data, and instructions needed to reproduce the main experimental results (either in the supplemental material or as a URL)?
    \answerYes{See Section 5, Appendix C, and supplemental material.}
  \item Did you specify all the training details (e.g., data splits, hyperparameters, how they were chosen)?
    \answerYes{See Section 3.5 and 4.2, and Appendix D.6.}
	\item Did you report error bars (e.g., with respect to the random seed after running experiments multiple times)?
    \answerYes{See Section 4.2.}
	\item Did you include the total amount of compute and the type of resources used (e.g., type of GPUs, internal cluster, or cloud provider)?
    \answerYes{See Appendix D.6.}
\end{enumerate}

\item If you are using existing assets (e.g., code, data, models) or curating/releasing new assets...
\begin{enumerate}
  \item If your work uses existing assets, did you cite the creators?
    \answerNA{}
  \item Did you mention the license of the assets?
    \answerNA{}
  \item Did you include any new assets either in the supplemental material or as a URL?
    \answerYes{See Section 5 and Appendix C.}
  \item Did you discuss whether and how consent was obtained from people whose data you're using/curating?
    \answerYes{See Appendix B.}
  \item Did you discuss whether the data you are using/curating contains personally identifiable information or offensive content?
    \answerYes{See Appendix B.}
\end{enumerate}

\item If you used crowdsourcing or conducted research with human subjects...
\begin{enumerate}
  \item Did you include the full text of instructions given to participants and screenshots, if applicable?
    \answerYes{See Section 3.4.}
  \item Did you describe any potential participant risks, with links to Institutional Review Board (IRB) approvals, if applicable?
    \answerYes{See Appendix B.}
  \item Did you include the estimated hourly wage paid to participants and the total amount spent on participant compensation?
    \answerYes{See Appendix D.3.}
\end{enumerate}

\end{enumerate}

%%%%%%%%%%%%%%%%%%%%%%%%%%%%%%%%%%%%%%%%%%%%%%%%%%%%%%%%%%%%

\newpage
\appendix
\renewcommand\thefigure{\Alph{section}\arabic{figure}}    
\renewcommand\thetable{\Alph{section}\arabic{table}}

% % \appendix

\section{Research Background}\label{background}

Various harmful memes propagate on Chinese platforms \citep{peng2019, zheng2016}.
While their creators and disseminators intend to simply express emotions or humor, and their original intent may be harmless, these memes have a significant negative impact on society when abused \citep{liu2016, lin2019}.
In this section, we individually explore the detrimental impacts of various harmful meme types, highlighting the importance of detecting these memes.

% 1
\textbf{Targeted Harmful.} 
Memes targeting specific individuals or groups can perpetuate hate speech, discrimination, and prejudice, fostering an environment of intolerance \citep{DBLP:conf/ijcai/SharmaAADMFHSN022}. 
They fuel online toxicity and create hostile environments.
Furthermore, they have the potential to incite real-world violence or harassment and worsen social divisions.  

\textbf{General Offensive.} Memes containing general offenses breed an aggressive culture prone to controversy and online violence \citep{zheng2016}.
Their influence extends beyond the individual, shaping the overall tone of the online environment. 
Additionally, such offensive content negatively impacts the development of correct values and healthy personalities among children \citep{yang2018}. 

\textbf{Sexual Innuendo.} 
While sexual innuendo content generally does not involve coercion in a sexual relationship, it can still be misconstrued due to gender and cultural differences, thereby contributing to sexualization \citep{thornhill1983human, koukounas2001psychological}. 
Moreover, inappropriate sexual innuendo may be considered sexual harassment \citep{otsri2020non}.

\textbf{Dispirited Culture.} 
Memes containing dispirited culture often evoke negative emotions and contribute to feelings of social isolation.
This leads to an increase in personal depression, making it difficult for individuals to have positive interactions and relationships with others \citep{LLDK202202018}.
Furthermore, the spread of these memes inadvertently undermines the value of positive thinking and promotes social anxiety  \citep{zhang2021}.

\section{Implementation Details}

\subsection{Details of Data Filtering}
% \tablename~\ref{Keyword} lists the search keywords for collecting memes, including several sensitive topics easily debated online as well as negative emotions and behaviors. 
% All the samples of \textsc{ToxiCN MM} are extracted from February 2023 to November 2023. 
For data filtering, we refer to the existing Chinese meme datasets \citep{DBLP:conf/nlpcc/LiLYXZ22, DBLP:conf/sigir/XuLZNZL022} and apply the following criteria:

\begin{itemize}[itemsep=0.5pt]
    \item The meme text must contain Chinese (including code-switching); memes only containing other languages are not allowed.
    \item The meme text must have actual semantics; 
    Thus, samples where the text is too brief, e.g. containing only modal particles, are removed.
    \item The meme must be readable. Hence, low-resolution samples that cannot be extracted inline text are excluded.
    \item The meme must be multimodal, meaning it should contain both the inline text and image information.
\end{itemize}

Here we present some examples of memes that were removed during the filtering process for failing to satisfy some of the above criteria, as shown in \figurename~\ref{filter_exp}. 
% In contrast,  reserving samples containing cartoons. 
% humorous and metaphor 

\setcounter{figure}{0}  
\begin{figure*}[b]
\centering
\includegraphics[width=14cm]{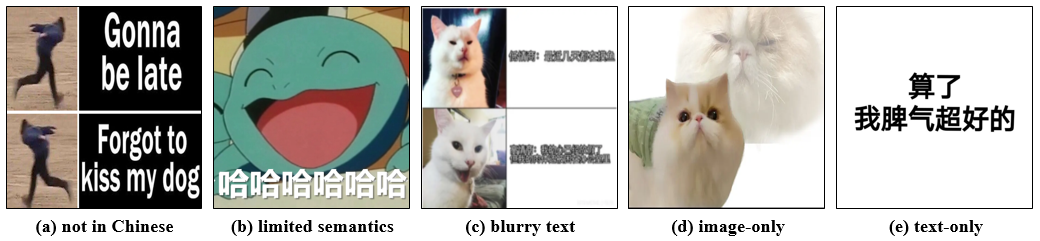}
% \vspace{-0.2in}
\caption{Examples of filtered memes and corresponding reasons. Among them, the inline text in (b) means "\textit{ha ha ha ha ha}", which is only an onomatopoeic word without semantics.}
% \vspace{-0.1in}
\label{filter_exp}
\end{figure*}

\subsection{Meme Containing Harmful Image}
\begin{figure}[htpb]
\centering
\includegraphics[width=6.5cm]{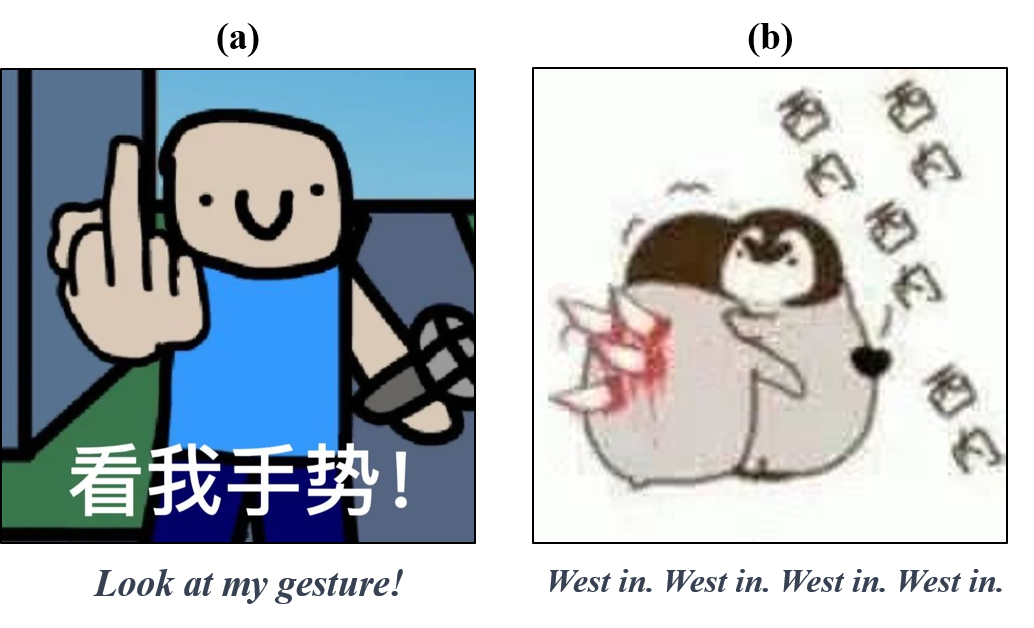}
\vspace{-0.075in}
\caption{Examples of harmful memes where images independently exhibit toxicity (\textit{Harm.I}).}
% \vspace{-0.025in}
\label{harmful_image}
\end{figure}

Based on the statistics listed in Table \ref{statistics}, the harmful meme where the image independently exhibits toxicity (\textit{Harm.I}) is sparse.
Here we present two samples to conduct a brief analysis, as shown in Figure \ref{harmful_image}.
Both examples contain general offensive content. In Exp. (a), the \textit{"middle finger"} is employed to convey aggression and contempt. 
In Exp. (b), both the inline text and image are independently harmful. The text "西内", literally translated as "\textit{west in}" in English, serves as a homonym for \textit{"go die"} in Japanese, while the image incorporates violent elements.

\subsection{Discussion of Annotation Consistency}\label{IAA}
% To mitigate the subjective biases in labeling, we guarantee the demographic diversity of annotators, as shown in \tablename~\ref{demographics}.  
After annotation, we calculate the Inter-Annotator Agreement (IAA) for each annotation hierarchy using Fleiss’ Kappa values. 
Among them, the stage with the highest disagreement pertains to determining whether a meme is harmful, with a Kappa value of 0.62, which is comparable to other harmful meme datasets like Harm-C (0.67) and Harm-P (0.68) \cite{DBLP:conf/emnlp/BlaierMW21}. 
Given the subtlety of harmful types in our dataset, this IAA is expected. 
These disagreements stem mainly from the humorous elements present in some samples of harmful memes, leading some annotators to consider them not toxic enough to classify them as harmful.
This reflects that subjective bias still influences the results of annotation to some extent, despite implementing several measures to mitigate biases.
In addition, the kappa values of discriminating harmful types and text-image combination characteristics are 0.73 and 0.86, respectively. 

\subsection{Statistics of Target Distribution}
During the annotation phase, we label specific targets of targeted harmful memes in \textsc{ToxiCN MM}.
The target distribution is shown in Table \ref{distribution}.

\begin{table}[htpb]
\small
  \centering
  \scalebox{0.9}{
\begin{tabular}{lm{1.75cm}<{\centering}m{1.5cm}<{\centering}}
    \toprule
    \textbf{Category} & \textbf{Num} & \textbf{Ratio}/\% \\
    \midrule
    Gender    & 417   & 41.04 \\
    Region    & 176   & 17.32 \\
    Occupation    & 71    & 6.99 \\
    Age    & 66    & 6.50 \\
    Body    & 55    & 5.41 \\
    Race    & 33    & 3.25 \\
    Individual    & 29    & 2.85 \\
    Health    & 27    & 2.66 \\
    LGBTQ+   & 22    & 2.17 \\
    Others    & 120   & 11.81 \\
    \midrule
    Total     & 1,016  & 100.00 \\
    \bottomrule
    \end{tabular}%
    }
    \vspace{0.1in}
  \caption{Target distribution of targeted harmful memes in \textsc{ToxiCN MM}.}
  % \vspace{-0.05in}
  \label{distribution}%
\end{table}%

\subsection{Experimental Details}\label{Appendix_exp} 

% \subsection{Hyperparameters Details}
In the evaluation phase, the \textit{harmful type identification} task is conducted as a five-classification, including \textit{non-harmful} and four harmful types.
% (i.e., targeted harmful memes, general offense, sexual innuendo, and dispirited culture).
The specific versions of each baseline are listed in Table B2.
To minimize experimental error, all experiments are repeated five times. 

For fine-tuned models, we acquire their original parameters from Hugging Face\footnote{\url{https://huggingface.co/}}.
Weighted cross-entropy is employed to tackle category imbalances, and AdamW is selected as the optimizer.
During the training phase, an early stopping mechanism is implemented to prevent overfitting. 
To reduce experimental error, all experiments are repeated five times with different random seeds. 
Details of the hyperparameter settings are presented in Table B3. 
All experiments are conducted using a GeForce RTX 3090 GPU.

% and Google\footnote{\url{https://ai.google.dev/gemini-api}} 

For GPT-3.5 and GPT-4, we employ the official API provided by Openai\footnote{\url{https://openai.com/}} to invoke them. 
We respectively design the instruction templates for the two tasks, shown in \tablename~\ref{template_1} and \tablename~\ref{template_2}.
In addition to the definition of Chinese harmful memes, specific evaluation criteria and steps are also provided in the template that adapt to our annotation process.
For example, in the evaluation criteria of the template for the task of \textit{harmful type identification}, we emphasize that the meme containing general offense does not have specific targets, and the sexual innuendo sample does not contain sexist or sexually assault content.
In future work, we plan to further optimize the design of the instruction template to improve the LLM detection performance of Chinese harmful memes. 
Furthermore, we will also evaluate the detection performance of the LLM in the few-shot scenario.

\vspace{0.15in}
\begin{minipage}{0.5\textwidth}
            \renewcommand{\arraystretch}{1.1}
		\centering
		\label{model_version}
		\setlength{\tabcolsep}{0.5 mm}
		\scalebox{0.85}{
           \begin{tabular}{cc}
            \midrule
            \textbf{Model} & \textbf{Version} \\
            \midrule
            RoBERTa & \texttt{chinese-roberta-wwm-ext-base} \\
            ResNet & \texttt{resnet-101} \\
            ViT   & \texttt{vit-base-patch16-224} \\
            CLIP  & \texttt{chinese-clip-vit-base-patch16} \\
            \midrule
            GPT-3.5 & \texttt{gpt-3.5-turbo} \\
            GPT4  & \texttt{gpt-4-vision-preview} \\
            % Gemini  & \texttt{gemini-1.0-pro-vision-latest} \\
            \bottomrule
            \end{tabular}%		
	}
	\captionof{table}{Specific version of each baseline.}
\end{minipage}
\begin{minipage}{0.5\textwidth}
            \renewcommand{\arraystretch}{1.2}
		\centering
		\label{hyp}
		\setlength{\tabcolsep}{0.5mm}
		\vspace{-1mm}
		\scalebox{0.925}{
		\begin{tabular}{m{3cm}<{\centering}m{2cm}<{\centering}m{1.5cm}<{\centering}}
        \toprule
        \textbf{Hyperparameters} & \textbf{ResNet/ViT} & \textbf{Others} \\
        \midrule
        epochs &20  &10  \\
        batch size &32  &32  \\
        learning rate &5e-5  & 1e-5 \\
        padding size & - & 80 \\
        dropout rate &0.1 & 0.1  \\
        % \midrule
        % $\lambda$ & 0.5 & 0.01 \\
        \bottomrule
        \end{tabular}
    	}
	\captionof{table}{Hyperparameters for experiments.}
\end{minipage}

\vspace{0.1in}

\begin{table*}[htpb]
\small
  \centering
    \begin{tabular}{p{45em}}
    \toprule
    \textbf{中文有害模因的定义:}\newline{}中文有害模因是由图像和中文行内文字组成的多模态单元，具有对个人、组织、社区、社会群体或整个社会造成伤害的潜力。它们往往反映和强化了中国互联网上潜在的负面价值观和文化态度。\newline{}
    \textbf{Definition of Chinese Harmful Meme}:\newline{}Chinese harmful memes are multimodal units consisting of an image and Chinese inline text that have the potential to cause harm to an individual, an organization, a community, a social group, or society as a whole. They often reflect and reinforce underlying negative values and cultural attitudes on the Chinese Internet.\\ 
    \midrule
    \textbf{\textit{Input for Harmful Meme Detection:}}\newline{}
    给定模因~<Image>，其内联文本为~<Text>~/~仅根据模因的内联文本~<Text>，将其分类为有害或无害。\newline{}
    评估标准:\newline{}<中文有害模因的定义>~根据定义，符合以下任何一个标准的模因将被判定为有害。\newline{}1. 模因表达了对特定个人或社会群体的厌恶、偏见或刻板印象。\newline{}2. 模因包含了讽刺或粗鲁的内容，没有特定的目标。\newline{}3. 模因暗示了性意图，以引起性冲动。\newline{}4. 模因包含了颓废和绝望的情绪，或传达了一种自我否定的态度。\newline{}
    评估步骤:\newline{}1. 结合图像和文本~/~仅根据文本分析模因的含义；\newline{}2. 将模因与评估标准进行比较，检查是否有任何匹配；\newline{}3. 只返回“有害”或“无害”的结论；\newline{}4. (可选)~提供做出决策的理由。
    \newline{}
    Given the meme <Image> and its inline text <Text>~/~Based solely on the inline text of a meme <Text>, classify it as harmful or harmless.
    \newline{}
    \textbf{Evaluation Criteria:}\newline{}<Definition of Chinese Harmful Meme>~According to the definition, meme that meets any of the following criteria will be judged as harmful.\newline{}1. The meme expresses disgust, prejudice, or stereotypes towards specific individuals or social groups.  \newline{}2. The meme encompasses sarcastic or rude content without specific targets.\newline{}3. The meme implies sexual intent to provoke sexual arousal.\newline{}4. The meme integrates decadent and desperate emotions or conveys a self-negative attitude. 
    \newline{}
    \textbf{Evaluation Steps:}\newline{}1. Analyze the meaning of meme by combining the inline text and image~/~solely based on the inline text;\newline{}2. Compare the meme against the criteria to check for any matches.\newline{}3. Solely return a conclusion of "\textit{harmful}" or "\textit{harmless}".\newline{}4. (Optional) Provide reasons for the decision. \\
    \bottomrule
    \end{tabular}%
  % \vspace{0.05in}
  \caption{Instruction templates for the task of \textit{harmful meme detection}.}
  \label{template_1}%
\end{table*}%

\begin{table*}[htpb]
\small
  \centering
    \begin{tabular}{p{45em}}
    \toprule
    \textbf{中文有害模因的定义:}\newline{}中文有害模因是由图像和中文行内文字组成的多模态单元，具有对个人、组织、社区、社会群体或整个社会造成伤害的潜力。它们往往反映和强化了中国互联网上潜在的负面价值观和文化态度。\newline{}
    \textbf{Definition of Chinese Harmful Meme}:\newline{}Chinese harmful memes are multimodal units consisting of an image and Chinese inline text that have the potential to cause harm to an individual, an organization, a community, a social group, or society as a whole. They often reflect and reinforce underlying negative values and cultural attitudes on the Chinese Internet.\\ 
    \midrule
    \textbf{\textit{Input for Harmful Type Identification}}\newline{}
    给定模因~<Image>，其内联文本为~<Text>~/~仅根据模因的内联文本~<Text>，将其分类为以下其中一类。\newline{}
    评估标准:\newline{}<中文有害模因的定义>~根据定义，如果模因具有以下特征之一，将其分类到相应的类别：\newline{}1. 如果模因表达了对特定个人或社会群体的厌恶、偏见或刻板印象，返回“A”。\newline{}2. 如果模因包含了讽刺或粗鲁的内容，但没有特定的目标，返回“B”。\newline{}3. 模因暗示了性意图，以引起性冲动，且不包含性别歧视或性侵犯的内容，返回“C”。\newline{}4. 如果模因包含了颓废和绝望的情绪，或传达了一种自我否定的态度，返回“D”。\newline{}3. 如果模因不属于以上任何的毒性类别或无法判断，返回“E”。\newline{}
    评估步骤:\newline{}1. 结合图像和文本~/~仅根据文本分析模因的含义；\newline{}2. 将模因与评估标准进行比较，检查是否有任何匹配；\newline{}3. 从"A"、"B"、"C"、"D"、"E"中选择对应的类别。
    \newline{}
    Given the meme <Image> and its inline text <Text>~/~Based solely on the inline text of a meme <Text>, classify it into one of the categories. 
    \newline{}
    \textbf{Evaluation Criteria:}\newline{}<Definition of Chinese Harmful Meme>~According to the definition, if the meme exhibits any of the following characteristics, it will be assigned to that corresponding category.\newline{}
    1. If the meme expresses disgust, prejudice, or stereotypes towards specific individuals or social groups, return "A". (\textit{Targeted Harmful})\newline{}
    2. If the meme encompasses sarcastic or rude content without specific targets, return "B". (\textit{General Offensive})\newline{}
    3. If the meme implies sexual intent to provoke sexual arousal, and does not contain sexist or sexually assault content, return "C". (\textit{sexual innuendo})\newline{}
    4. If the meme integrates decadent and desperate emotions or conveys a self-negative attitude, return "D". (\textit{Dispirited Culture}) \newline{} 
    5. If the meme does not fall into any of the above harmful types or cannot be determined, return "E". (\textit{Non-Harmful})
    \newline{}
    \textbf{Evaluation Steps:}\newline{}1. Analyze the meaning of meme by combining the inline text and image~/~solely based on the inline text;\newline{}2. Compare the meme against the criteria to check for any matches.\newline{}3. Choose the corresponding category from A/B/C/D/E.\\
    \bottomrule
    \end{tabular}%
  % \vspace{-0.05in}
  \caption{Instruction templates for the task of \textit{harmful type identification}.}
  \label{template_2}%
\end{table*}%

\section{Supplementary Experiments}

\subsection{Performance on Diverse Modality Combination}

\setcounter{figure}{0}  
\begin{wrapfigure}{r}{6.75cm}
\vspace{-0.05in}
\centering
\includegraphics[width=6.75cm]{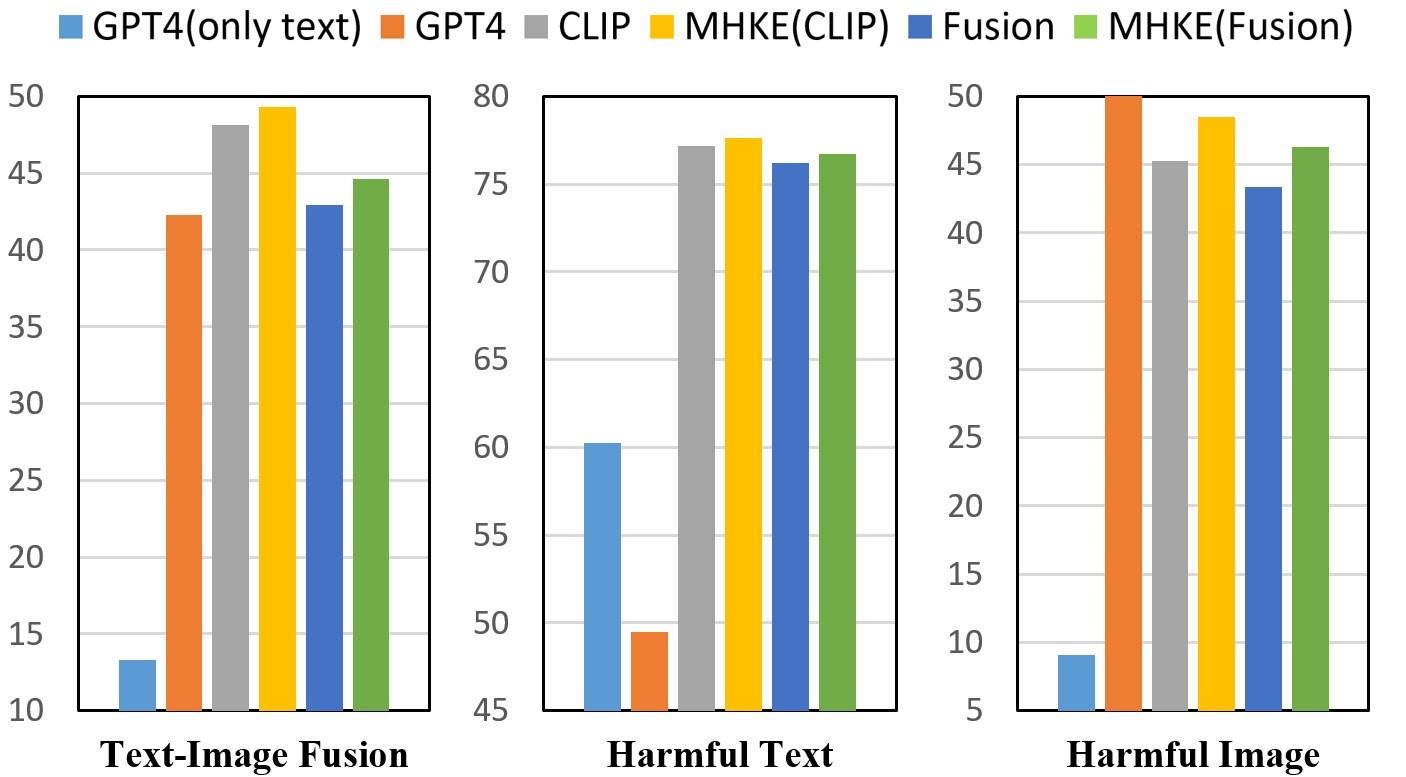}
% \vspace{-0.25in}
\caption{Accuracy of Chinese harmful memes in the test set of \textsc{ToxiCN MM} with different modality combination features.}
% \vspace{-0.05in}
\label{fusion}
\end{wrapfigure}

For a supplementary analysis, we evaluate the detection performance of harmful memes with different modality combination features, as shown in Figure \ref{fusion}.

Compared to other combinations, memes containing harmful text are more likely to be successfully detected by the models, especially PLMs.
This is because PLMs can effectively learn the unique expressions of the Chinese language during the fine-tuning stage.
In contrast, for memes containing harmful images, GPT-4 demonstrates stronger performance than fine-tuned PLMs, with an average increase of 6.5\%, illustrating its ability to effectively review the content of input images.
Additionally, we note that GPT-4's performance on text-image fusion is comparable to that of PLMs, showcasing its capability to understand and reason with information that combines both textual and visual elements effectively.
Meanwhile, compared to the model incorporating only inline text, i.e., GPT-4 (only text), GPT-4 shows a 12.7\% decrease in performance for memes containing harmful text.
This also supports the conclusion that benign images relatively affect the detection of GPT-4 in Chinese harmful memes.
Additionally, after introducing MKE, models show improvement in detecting Chinese harmful memes with different modality combinations.

\subsection{Evaluation of Chinese MLLMs}\label{CN_llms}

\setcounter{table}{0}  
\begin{wraptable}{r}{6.5cm}
\renewcommand{\arraystretch}{1.2}
\small
  \centering
  \scalebox{0.875}{
    \begin{tabular}{lm{0.7cm}<{\centering}m{0.7cm}<{\centering}m{0.7cm}<{\centering}}
    \toprule
    Model & P & R & F1 \\
    \midrule
    % Ziya-Visual (Ziya-LLaMA-13B) &       &       &  \\
    VisualGLM (ChatGLM-6B) &  57.68   &  56.94     & 57.36 \\
    Qwen-VL (Qwen-7B) & 59.53   &  58.74     & 58.87 \\
    VisCPM (CPMBee-10B) & 54.32      & 54.63   & 54.47 \\
    \bottomrule
    \end{tabular}%
    }
    \caption{The performance of several Chinese MLLMs on the \textit{harmful meme detection} task.}
    % \vspace{-0.05in}
  \label{cn_mllm}%
\end{wraptable}%

In this section, we utilize \textsc{ToxiCN MM} to evaluate the performance of Chinese LLMs in detecting Chinese harmful memes under the zero-shot scenario.
Due to the unavailability of APIs for multimodal dialogue in existing commercial Chinese LLMs (e.g., Wenxin Yiyan), we only evaluate the effect of several open-source Chinese LLMs,
including VisualGLM \citep{du2022glm}, Qwen-VL \citep{Qwen-VL}, and VisCPM 
 \citep{DBLP:journals/corr/abs-2308-12038}. 
The results are shown in Table \ref{cn_mllm}.
 
Based on the result, we note that the detection performance of these open-source Chinese LLMs is not satisfactory compared to other baselines shown in Table \ref{result}.
This is because these models have weak multimodal reasoning capabilities and contain limited background knowledge required for detection.
% In future work, we will further employ instruction fine-tuning to enhance the effect of models, and continuously evaluate the performance of state-of-the-art LLMs to ensure the effectiveness of \textsc{ToxiCN MM}.

\section{Licensing and Maintenance Plan}

\subsection{Licensing}
We confirm that the dataset is licensed under the Creative Commons Attribution-NonCommercial 4.0 International (CC BY-NC 4.0) license. 
% This means that others are free to share and adapt the dataset for non-commercial purposes, provided appropriate credit is given, and any derivatives are also non-commercial.

\subsection{Maintenance Plan}

We will regularly update the \textsc{ToxiCN MM} dataset by adding new samples collected from Chinese social platforms. 
These updates, scheduled annually, aim to maintain the dataset's currency and comprehensiveness. 
Additionally, we will open the \textit{Community} section of \texttt{Huggingface} for community contributions, subjecting all submissions to rigorous review to ensure alignment with the dataset's quality standards. 
Moreover, based on the feedback, we will periodically enhance annotation guidelines and expand annotation types to provide a more detailed and diverse dataset. Future updates will also include enriched metadata to offer better context and support more analyses.

To address any concerns or queries related to the dataset, we will establish a dedicated support team reachable via email or through a support portal on the website. 
Furthermore, an issue tracking system will be implemented to document and monitor reported issues, enabling users to report bugs and suggest improvements. 
We encourage user feedback through a structured feedback loop, which will inform regular audits aimed at maintaining dataset integrity and quality. Periodic transparency reports will be published to keep users informed about encountered issues, resolutions, and overall dataset improvements, fostering trust and transparency within the user community.

\clearpage

\clearpage\end{CJK*}

\end{document}